\documentclass[10pt,twocolumn,letterpaper]{article}

\usepackage[pagenumbers]{cvpr} 
\usepackage[accsupp]{axessibility}
\usepackage{multirow}
\usepackage{balance}
\usepackage{fdsymbol}
\usepackage{arydshln}
\usepackage{graphicx}

\def\symbola{\textcolor[rgb]{1,0.7,0}{$\vardiamondsuit$}}
\def\symbolb{\textcolor[rgb]{0.,0.5,0.9}{$\vardiamondsuit$}}
\def\symbolc{\textcolor[rgb]{0.9,0.3,0.4}{$\vardiamondsuit$}}
\def\symbold{\textcolor[rgb]{0.8,0.4,0.9}{$\vardiamondsuit$}}

\usepackage[dvipsnames]{xcolor}

\definecolor{cvprblue}{rgb}{0.21,0.49,0.74}
\usepackage[pagebackref,breaklinks,colorlinks,citecolor=cvprblue]{hyperref}

\def\paperID{8784} 
\def\confName{CVPR}
\def\confYear{2024}

\title{Adapting to Length Shift: FlexiLength Network for Trajectory Prediction}

\author{Yi Xu \quad Yun Fu\\
Northeastern University, USA \\
{\tt\small xu.yi@northeastern.edu, yunfu@ece.neu.edu}
}

\begin{document}
\maketitle
\begin{abstract}
Trajectory prediction plays an important role in various applications, including autonomous driving, robotics, and scene understanding. Existing approaches mainly focus on developing compact neural networks to increase prediction precision on public datasets, typically employing a standardized input duration. However, a notable issue arises when these models are evaluated with varying observation lengths, leading to a significant performance drop, a phenomenon we term the Observation Length Shift. To address this issue, we introduce a general and effective framework, the FlexiLength Network (FLN), to enhance the robustness of existing trajectory prediction techniques against varying observation periods. Specifically, FLN integrates trajectory data with diverse observation lengths, incorporates FlexiLength Calibration (FLC) to acquire temporal invariant representations, and employs FlexiLength Adaptation (FLA) to further refine these representations for more accurate future trajectory predictions. Comprehensive experiments on multiple datasets, \ie, ETH/UCY, nuScenes, and Argoverse 1, demonstrate the effectiveness and flexibility of our proposed FLN framework.
\end{abstract}    
\section{Introduction}
\label{sec:intro}
The goal of trajectory prediction is to predict the future locations of agents conditioned on their past observed states, a key step in understanding their motion and behavior patterns. This task is essential yet challenging in many real-world applications such as autonomous driving~\cite{zhu2021simultaneous, wang2022ltp, hazard2022importance, hu2023planning}, robotics~\cite{cheng2020towards, mahdavian2023stpotr}, and scene understanding~\cite{chi2023adamsformer, xu2023human}. Recent advancements in deep learning have significantly improved the accuracy of trajectory predictions~\cite{xu2020cf, xu2021tra2tra, rowe2023fjmp, mao2023leapfrog, jiang2023motiondiffuser, chen2023unsupervised, zhou2023query, gu2023vip3d, aydemir2023adapt, bae2023eigentrajectory, shi2023trajectory, chen2023traj, seff2023motionlm, park2023leveraging}. However, achieving such promising prediction performance typically requires complex models and extensive computational resources. Additionally, many state-of-the-art methods are developed on public datasets~\cite{zhan2019interaction, chang2019argoverse, sun2020scalability, wilson2023argoverse} with a fixed observation length, failing to consider the potential discrepancies between training conditions and diverse real-world scenarios. Consequently, the fixed training setups are often too rigid to meet in practice, especially when faced with environmental changes or observational variables.

\begin{figure}[t]
  \centering
   \includegraphics[width=0.98\linewidth]{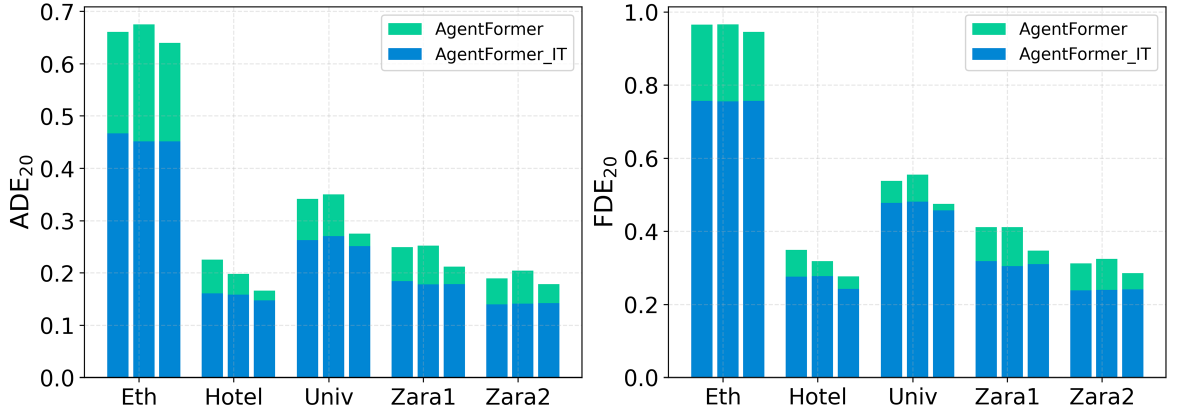}
   \caption{The Observation Length Shift phenomenon is a common issue in the trajectory prediction task. The AgentFormer model is trained with a standard observation length of 8 timesteps and tested at varying lengths to compare with Isolated Training (IT). For each dataset, the bar groups from left to right correspond to observation lengths of 2, 4, and 6 timesteps, respectively.}
   \label{fig:intro}
\end{figure}

Recently, some studies have explored the issue of training-testing discrepancy in trajectory prediction. For example, work~\cite{xu2023uncovering} proposes a solution for scenarios where observations might be incomplete due to occlusions or limited viewpoints, by developing a method that combines trajectory imputation and prediction. Other research efforts~\cite{xu2022adaptive, bahari2022vehicle, ivanovic2022expanding, huynh2023online} focus on tackling the domain shift challenge, where the testing environment differs from the training. Additionally, studies~\cite{sun2022human, monti2022many, li2023bcdiff} address situations with limited observations, such as having only one or two data points available. These approaches try to mitigate the training-testing gap from various angles. Acknowledging these efforts, the importance of addressing the training-testing gap is highlighted. In our paper, we concentrate on the less-explored issue of observation length discrepancy.

For certain well-known architectures such as RNN~\cite{hochreiter1997long} and Transformer~\cite{vaswani2017attention}, the ability to process inputs of varying lengths presents a straightforward solution: train the model with the standard input length and then evaluate at different input lengths. This solution is contrasted with Isolated Training (IT), where the model is trained at the specific observation length and evaluated with the same input length. We conducted experiments using a Transformer-based method AgentFormer~\cite{yuan2021agentformer} on dataset ETH/UCY~\cite{pellegrini2009You, lerner2007crowds} with results illustrated in \cref{fig:intro}. The results reveal a clear performance discrepancy between these two solutions, suggesting performance declines when not evaluated at the training length. We term this phenomenon the Observation Length Shift. It should be noted that while the performance of IT marginally improves, it requires multiple training rounds for different observation lengths. This observation raises two crucial questions:
\begin{enumerate}
    \item \textbf{What are the underlying reasons behind this performance discrepancy?}
    \item \textbf{Can this issue be resolved without substantial modifications or increased computational expense?}
\end{enumerate}

Prompted by these questions and given the popularity of Transformer-based models in the trajectory prediction task, we conducted an in-depth analysis of a Transformer-based trajectory prediction method, as detailed in \cref{sec:look}. This investigation revealed two main causes, which we summarized as positional encoding deviation and normalization shift. Building on these findings, we introduce a robust and adaptable framework, the FlexiLength Network (FLN), to overcome the Observation Length Shift problem. FLN effectively merges trajectory data from various observation lengths, integrates FlexiLength Calibration (FLC) to create time-invariant representations, and applies FlexiLength Adaptation (FLA) to enhance these representations, thereby improving the trajectory prediction precision. Notably, FLN requires only one-time training, without substantial modifications, but offers the flexibility to different observation lengths, thus addressing our second question.

Despite some studies focusing on positional encoding deviation~\cite{dehghani2018universal, nam2020bert} and normalization shift~\cite{yu2018slimmable, yu2019universally}, our work addresses the Observation Length Shift, a commonly overlooked issue in trajectory prediction. We propose the FlexiLength Network (FLN) as a practical and non-trivial solution, especially valuable in real-world scenarios. Unlike Isolated Training (IT), which is time and memory-expensive for adapting to various observation lengths, FLN offers a more resource-efficient alternative. Even though modern architectures accommodate various input lengths, they often suffer performance discrepancies, a gap that FLN effectively bridges while preserving the original model design. Furthermore, FLN can be applied to nearly all existing models that are based on the Transformer architecture. The key contributions of our work are organized as follows:
\begin{itemize}
    \item We identify and thoroughly analyze the Observation Length Shift phenomenon in trajectory prediction, pinpointing the factors that lead to performance degradation and guiding the direction of our solution.
    \item We introduce the FlexiLength Framework (FLN), a robust and efficient framework requiring only one-time training. It includes FlexiLength Calibration (FLC) and FlexiLength Adaptation (FLA) for learning temporal invariant representations, boosting prediction performance with different observation lengths.
    \item We validate the effectiveness and generality of FLN through comprehensive experiments on multiple datasets. It demonstrates consistent superiority over Isolated Training (IT) with various observation lengths.
\end{itemize}

\section{Related Work}
\label{sec:related}

\noindent
\textbf{Trajectory Prediction.} The objective is to forecast the future trajectories of agents based on their historical observations. The primary focus has been on understanding the social dynamics among these agents, leading to the development of various methodologies, utilizing Graph Neural Networks (GNNs) as seen in works like~\cite{kosaraju2019social, sun2020recursive, mohamed2020social, shi2021sgcn, xu2022groupnet, li2022graph, bae2022learning}. The inherently uncertain and diverse nature of future trajectories has led to the development with generative models, including Generative Adversarial Networks (GANs)~\cite{gupta2018social, sadeghian2019sophie, li2019conditional, amirian2019social, kosaraju2019social}, Conditional Variational Autoencoders (CVAEs)~\cite{mangalam2020It, xu2022remember, ivanovic2019trajectron, salzmann2020trajectronplusplus, xu2022socialvae}, and Diffusion models~\cite{gu2022stochastic, mao2023leapfrog, jiang2023motiondiffuser}. These models have achieved promising results on various datasets. However, most of these methods prioritize prediction accuracy while overlooking the discrepancies that often arise in real-world applications, resulting in decreased adaptability under varying conditions. In contrast, our work focuses on the variation in observation lengths, a challenge that many existing methods face. It is noteworthy that some methods based on RNN~\cite{ivanovic2019trajectron, salzmann2020trajectronplusplus, chiara2022goal} and Transformer~\cite{yuan2021agentformer, tsao2022social, giuliari2020transformer, yu2020spatio} can handle varying input lengths. However, these approaches show limitations under observation length shifts. Our work delves into this specific challenge, proposing a comprehensive solution that not only enhances the robustness of existing methods but also improves their performance on varying observation lengths.


\noindent
\textbf{Training-Testing Discrepancy.} The gap between training and testing can present in different ways, either in input types or environmental conditions. Recent studies have begun to uncover various aspects of discrepancy in trajectory prediction. Some methods~\cite{xu2022adaptive, ivanovic2022expanding, bahari2022vehicle, huynh2023online} have focused on the environmental shifts between training and testing, offering innovative solutions. Another approach~\cite{wang2023enhancing} deals with the absence of High-Definition (HD) maps during testing, employing Knowledge Distillation (KD) to transfer map knowledge from a more informed teacher model to a student model. Another aspect recently explored is the completeness of observations. Studies like~\cite{sun2022human, monti2022many, li2023bcdiff} assume only limited observed trajectory data, specifically one or two frames, is accessible for forecasting. This assumption has led to the concept of instantaneous trajectory prediction. Other studies~\cite{xu2023uncovering, lange2023scene} have addressed the challenges posed by incomplete observations due to occlusions or viewing limitations in real environments. Another recent work~\cite{park2024improving} attempts to diminish internal discrepancies across datasets collected in arbitrary temporal configurations. While these methods have recognized various aspects of training-testing discrepancies, the issue of observation length shift remains relatively unexplored. Our proposed framework stands out in its generality and effectiveness, improving the ability to handle trajectory data across different observation lengths.

\noindent
\textbf{Test-Time Adaptation.} The objective of test-time adaptation is to adapt models to new data during testing. A common approach includes integrating an auxiliary task~\cite{dosovitskiy2014discriminative, noroozi2016unsupervised, gidaris2018unsupervised} that employs straightforward self-supervision techniques on the target domain, which typically requires modifications to the training process in the source domain to include this auxiliary task. However, recent strategies~\cite{wang2021tent, liu2021ttt++} enable straightforward adjustments in the target domain, avoiding the need to modify the training process. Unlike these methods, our proposed framework only requires one-time training, eliminating the necessity for adjustments during training or additional tuning during inference. Moreover, our framework functions independently, without the need for evaluation data during the training phase.
\section{A Close Look at Performance Degradation}
\label{sec:look}

\begin{figure}[t]
  \centering
   \includegraphics[width=0.98\linewidth]{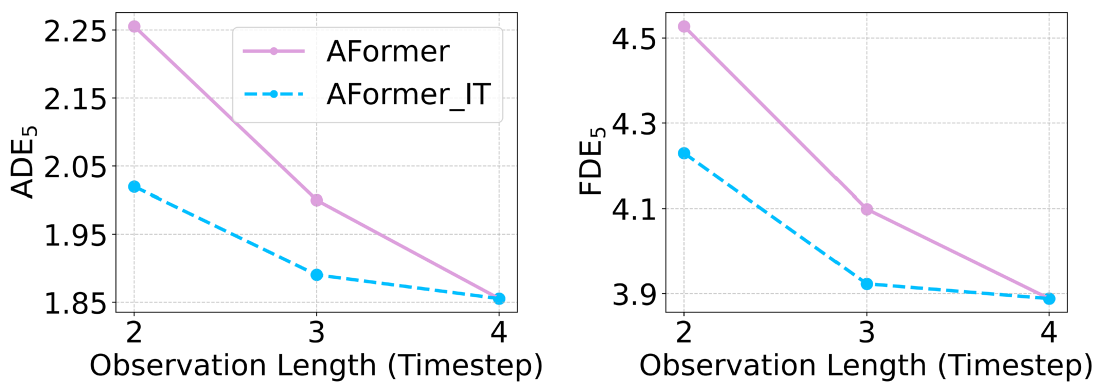}
   \caption{ADE$_{5}$ and FDE$_{5}$ results for the AgentFormer model, which is trained on the nuScenes dataset using a standard observation length of 4 timesteps, and tested at shorter 
 observation lengths of 2 and 3 timesteps. These results are compared to those obtained through Isolated Training (IT).}
   \label{fig:aformer}
\end{figure}

\noindent
\textbf{Observation Length Shift.}
The Observation Length Shift phenomenon becomes apparent when models, trained with a specific observation length, are tested using varying lengths, as illustrated in \cref{fig:intro}. To investigate further, we train the AgentFormer model~\cite{yuan2021agentformer} using the nuScenes dataset~\cite{caesar2020nuscenes} and evaluate its performance across different observation lengths, with the results presented in \cref{fig:aformer}. These results reveal noticeable performance gaps when the model is evaluated with different observation lengths compared to Isolated Training. Notably, the performance degradation is smaller when the testing observation length is close to the training observation length. 
\begin{figure}[t]
  \centering
\includegraphics[width=0.98\linewidth]{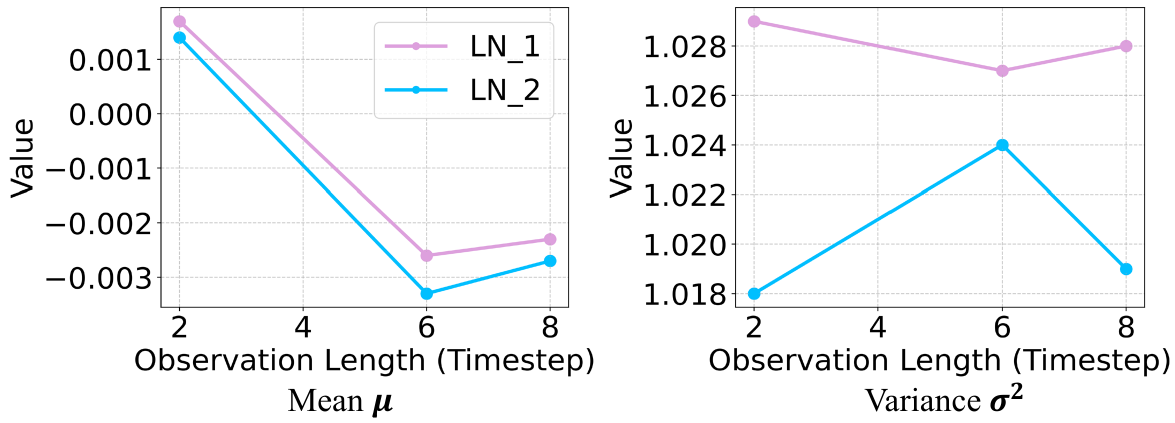}
   \caption{Layer Normalization statistics in two different layers of the Transformer encoder within the AgentFormer model, isolatedly trained on the Eth dataset at observation lengths of 2, 6, and 8 timesteps.}
   \label{fig:mean_var}
\end{figure}

\begin{figure*}[t]
  \centering
   \includegraphics[width=0.98\linewidth]{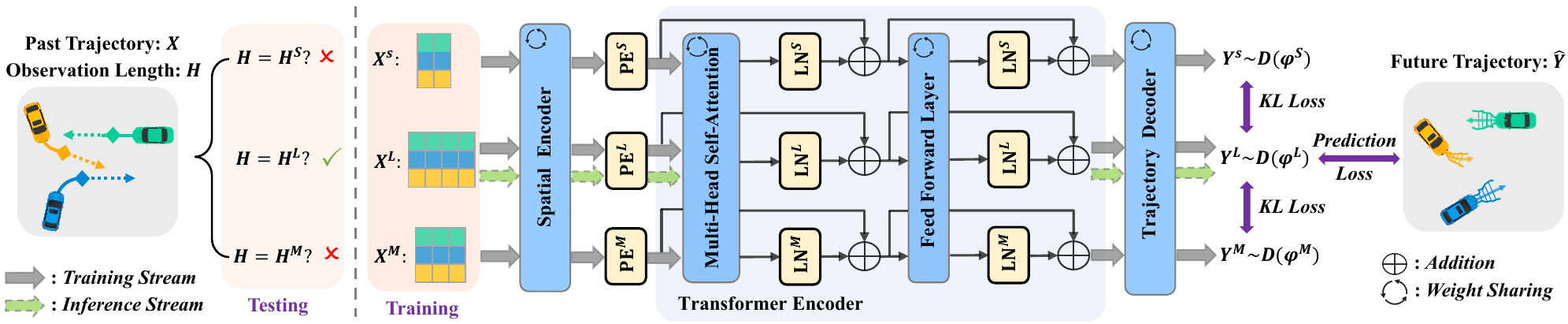}
   \caption{Illustration of our FlexiLength Network (FLN). The map encoding branch is omitted for simplicity. During training, with inputs of varying observation lengths $H^{S}$, $H^{M}$, and $H^{L}$, we utilize FlexiLength Calibration (FLC) to acquire temporal invariant representations. Furthermore, FlexiLength Adaptation (FLA) is employed to align these invariant representations with different sub-networks, thereby augmenting the model capabilities. During inference, the sub-network with the closest match in observation length is activated.}
   \label{fig:fra}
\end{figure*}

\noindent
\textbf{Positional Encoding Deviation.}
Previous studies~\cite{nam2019number, hahn2020theoretical} have demonstrated that Transformer models struggle to generalize to input lengths they haven't encountered before, due to a lack of inductive bias. Other works~\cite {dehghani2018universal, nam2020bert} identify positional encoding as a critical factor that limits the generalization ability to new lengths and suggest model structure adjustments to enhance the positional encoding. Focusing on the Transformer-based models, we analyze the positional encoding in the AgentFormer. The positional encoding feature $\tau_n^{t}(k)$ for agent $n$ at timestep $t$ is calculated as follows:
\begin{equation}
    \label{eq:1}
    \tau_n^{t}(k) = \left\{
    \begin{aligned}
    & \sin((t + H)/{10000^{k/d_t}}), k \text{ is even} \\
    & \cos((t + H)/{10000^{(k-1)d_t}}), k \text{ is odd}
    \end{aligned}
    \right. ,
\end{equation}
where $k$ represents the feature index, $d_t$ is the feature dimension, and $H$ is the observation length. It is obvious that simply removing observations without compensation will cause the length change, subsequently altering the positional encoding.
Consequently, we hypothesize that the positional encoding deviation is one crucial factor to the performance gap. Although other Transformer-based models~\cite{zhou2022hivt, zhou2023query} use a fully learnable positional embedding, our experiments in \cref{sec:exp} indicate that this type of positional embedding remains a key reason.

\noindent
\textbf{Normalization Shift.}
Previous works~\cite{yu2018slimmable, yu2019universally} have studied the issue of normalization shift in the context of image and video classification. They found that altering network widths impacts Batch Normalization (BN), leading to variations in the means and variances of aggregated features and thereby affecting feature aggregation consistency. Given that Layer Normalization (LN) is widely used in Transformer-based models and operates similarly to BN, we hypothesize that the discrepancy in LN statistics could be another factor affecting the generalization ability to different lengths. For example, if a model is trained using an observation length of $H$ and later evaluated using a length of $L$, the input to the LN would be the intermediate feature $f^{L}$ during testing, resulting in the following output:
\begin{equation}
    \label{eq:2}
    y^{L}=\gamma^{H}\frac{f^{L}-\mu^{H}}{\sqrt{{\sigma^{H}}^{2}+\epsilon}} + \beta^{H},
\end{equation}
where $\mu^{H}$, ${\sigma^{H}}^{2}$ are derived from the data distribution with input length $H$, and $\gamma^{H}$, $\beta^{H}$ are parameters learned during training with the same input length. We train the AgentFormer model using observation lengths $H$ and $L$ and then analyze the feature statistics in the Transformer encoder's two Layer Normalizations (LNs). Clear statistical differences at varying observation lengths can be observed in \cref{fig:mean_var}. Importantly, both $\mu$ and $\sigma$ are data-dependent, indicating that the differences arise directly from the data itself. Therefore, we believe that the variations in LN statistics are another reason for the Observation Length Shift phenomenon in Transformer-based models.
\section{FlexiLength Network}
\label{sec:method}
\subsection{Problem Formulation}
Despite some variations in terminology, the fundamental problem setup for multi-agent trajectory prediction remains the same across existing methods. Essentially, the objective is to predict the future trajectories of multiple agents, $\mathbf{Y} \in \mathbb{R}^{N \times T \times F}$, based on their past observed states $\mathbf{X}\in \mathbb{R}^{N \times H \times F}$. Here, $N$ represents the number of agents, $H$ is the length of observation, $T$ is the prediction duration and $F$ typically stands for 2, denoting coordinate dimensions. Common approaches aim to develop a generative model, denoted as $p_{\theta}(\mathbf{Y}|\mathbf{X}, \mathbf{I})$, where $\mathbf{I}$ represents the contextual information such as High-definition (HD) maps.

As previously mentioned, these methods often face challenges when there is a variation in observation lengths during evaluation. Hence, our objective is to develop a model $p_{\theta}(\mathbf{Y}|\mathbf{X}, \mathbf{I})$ that can be evaluated across a range of input lengths $\mathbf{H}$, and achieves similar or even better performance compared to Isolated Training (IT), where $\mathbf{H}$ denotes a set of different observation lengths.

\subsection{FlexiLength Calibration}
As observed in \cref{sec:look}, the performance gap is smaller when the evaluation length is closer to the training length. This observation inspired us to incorporate observations of varying lengths during training and to develop corresponding sub-networks for feature extraction from these sequences. Given an observation trajectory sequence $\mathbf{X}$, we can generate sequences of varying lengths either through truncation or a sliding window approach. In this way, we can collect three sequences: $X^{S}$, $X^{M}$, and $X^{L}$, corresponding to Short, Medium, and Long lengths, with corresponding lengths $H^{S}$, $H^{M}$, and $H^{L}$. During training, these three sequences are each fed into their corresponding sub-networks $F^{S}(\cdot)$, $F^{M}(\cdot)$, and $F^{L}(\cdot)$, for processing. 

\cref{fig:fra} showcases our FlexiLength Network (FLN) developed atop the Transformer-based architecture, where the map branch is omitted for simplicity. We categorize the components of typical Transformer-based models into several key parts: a Spatial Encoder for spatial feature extraction, a Positional Encoder (PE) for embedding positional information, a Transformer Encoder for temporal dependency modeling, and a Trajectory Decoder for generating trajectories. During training, the three inputs $X^{S}$, $X^{M}$, and $X^{L}$ are each processed by their respective sub-networks $F^{S}(\cdot)$, $F^{M}(\cdot)$, and $F^{L}(\cdot)$. During inference, the sub-network matching the observation length is exclusively activated. This approach sets up a computational stream in the FLN framework, enabling effective evaluation across various observation lengths.

\noindent
\textbf{Sub-Network Weight Sharing.} Given the three trajectory samples $X^{S}$, $X^{M}$, and $X^{L}$, we employ shared weights for the spatial encoder, temporal encoder, and trajectory decoder across all sub-networks, to find a group of parameters $\theta$ to extract the spatio-temporal features as follows:
\begin{equation}
    \label{eq:sub}
    Y^{*} \sim \mathcal{D}(\psi^{*}), \quad \mathcal{D}(\psi^{*}) =  F^{*}(X^{*}; \theta),
\end{equation}
where $* \in \{S, M, L\}$, and $\mathcal{D}(\psi^{*})$ denotes the distribution parameterized by $\psi^{*}$, that $Y^{*}$ adheres to, such as a bivariate Gaussian or Laplace mixture distribution~\cite{zhou2023query}.

This design contrasts with Isolated Training (IT) by being more efficient in terms of parameter usage, as it maintains only a single set of weights applicable to various lengths. Additionally, this shared-weight strategy enhances performance by implicitly providing the model with the prior knowledge that these three observed sequences are part of the same trajectory. This insight makes the model more resilient to observation length shifts.

\noindent
\textbf{Temporal Distillation.} 
Trajectory prediction models generally perform better with longer observations $X^{L}$ as more movement information is contained. Consequently, we regard $Y^{L}\sim \mathcal{D}(\psi^{L})$ as the most '\textit{accurate}' among the three predictions. To update the parameters of $F^{L}(\cdot)$, we employ a negative log-likelihood loss, represented as follows:
\begin{equation}
\label{eq:regloss}
    \mathcal{L}_{reg} = -\log(\mathbb{P}(\hat{Y}|\psi^{L})), 
\end{equation}
where $\hat{Y}$ is the ground truth trajectory. Note that as multiple agents always exist within a single trajectory, the above equation is a simplified version and omits the calculation for the average value per agent at each time step.

For updating the parameters of the other two sub-networks, $F^{S}(\cdot)$ and $F^{M}(\cdot)$, a direct approach might involve calculating a corresponding negative log-likelihood loss as shown in \cref{eq:regloss}. However, it has potential drawbacks. Since the weights are shared across all three sub-networks, the optimal parameters for $F^{L}(\cdot)$ may not be as effective for $F^{S}(\cdot)$ or $F^{M}(\cdot)$. Additionally, optimizing the log-likelihood loss for $\psi^{S}$ and $\psi^{M}$ could also hinder the performance of $F^{L}(\cdot)$. This is because their inputs $X^{S}$ and $X^{M}$, with less movement information, might lead to a poorer fit. To address this issue, we employ the KL divergence loss~\cite{kullback1997information} to calibrate and incorporate $\psi^{S}$ and $\psi^{M}$ into the computational process as follows: 
\begin{equation}
\label{eq:klloss}
    \mathcal{L}_{kl} = \text{KL}(\mathcal{D}(\psi^{L})||\mathcal{D}(\psi^{M})) + \text{KL}(\mathcal{D}(\psi^{L})||\mathcal{D}(\psi^{S})).
\end{equation}
Updating the weights of shared sub-networks through \cref{eq:klloss} aims to align the predicted distributions of the 'student' networks ($\mathcal{D}(\psi^{M})$ and $\mathcal{D}(\psi^{S})$) closely with the predicted distribution of the 'teacher' network ($\mathcal{D}(\psi^{L})$). This process facilitates the transfer of valuable knowledge from $F^{L}(\cdot)$ to both $F^{M}(\cdot)$ and $F^{S}(\cdot)$. The entire set of parameters for our FLN is updated as follows:
\begin{equation}
    \mathcal{L}_{FLN} = \mathcal{L}_{reg} + \lambda_{1}\cdot \mathcal{L}_{kl},
\end{equation}
where $\lambda$ serves to balance these two terms during backpropagation, we have set $\lambda = 1$ in our implementation.

Considering both Sub-Network Weight Sharing and Temporal Distillation, $\mathcal{L}_{reg}$ offers guidance for the prediction task, while $\mathcal{L}_{kl}$ offers intra-trajectory knowledge to network training. This will promote $\mathcal{D}(\psi^{L})$, $\mathcal{D}(\psi^{M})$, and $\mathcal{D}(\psi^{S})$ exhibit high similarity, as the length shift does not alter the distribution of one specific trajectory. Through this approach, FLN not only learns temporal invariant representations but also ensures ease of implementation, as it does not require any changes to the feature extractors.

\subsection{FlexiLength Adaptation}
We developed FLC to guide our FLN in learning temporal invariant representations. Additionally, we introduce FlexiLength Adaptation (FLA) to optimize the fit of these invariant features across different sub-networks, thereby further improving their representational capacity.

\noindent
\textbf{Independent Positional Encoding.} 
Our analysis in \cref{sec:look} identifies positional encoding as one factor that can confuse the model regarding observation length. To counter this, we implement independent positional encoding for each sub-network. Taking the positional encoding from the AgentFormer~\cite{yuan2021agentformer} as an example, we define the positional encoding for inputs with diverse observation lengths as follows:
\begin{equation}
    \label{eq:posti}
    \tau_n^{t}(k) = \left\{
    \begin{aligned}
    & \sin((t + H^{*})/{10000^{k/d_t}}), k \text{ is even} \\
    & \cos((t + H^{*})/{10000^{(k-1)d_t}}), k \text{ is odd}
    \end{aligned}
    \right. ,
\end{equation}
where $*\in{L, M, S}$. While recent Transformer-based models~\cite{zhou2022hivt, zhou2023query} have shifted to a fully learnable positional embedding instead of the traditional sinusoidal pattern,  we adopt this learnable positional embedding across each of our sub-networks. As a result, for each input sequence $X^{S}$, $X^{M}$, and $X^{L}$, we define unique learnable positional encoding $\tau^{S}$, $\tau^{M}$, and $\tau^{L}$. Note that the implementation of independent positional encoding for each sub-network is a resource-efficient solution, introducing only a negligible increase in parameters and computation.

\begin{table*}[t]
	\centering
   \scalebox{0.75}{
    \begin{tabular}{lccccccccc}
    \toprule
     \multicolumn{1}{l}{\multirow{2}{*}{Method}} 
     & \multicolumn{1}{c}{\multirow{2}{*}{Note}}
     & \multicolumn{1}{c}{\multirow{2}{*}{\#Param}}
     & \multicolumn{3}{c}{ADE$_{5}$/FDE$_{5}$ $\downarrow$ \quad K = 5 Samples}
     & \multicolumn{3}{c}{ADE$_{10}$/FDE$_{10}$ $\downarrow$ \quad K = 10 Samples} \\
    \cmidrule(lr){4-9}
     & & & 2 Timesteps & 3 Timesteps & 4 Timesteps & 2 Timesteps & 3 Timesteps & 4 Timesteps \\
    \midrule
    AFormer~\cite{yuan2021agentformer} & - & 6.67M 
    & 2.26/4.53 & 2.00/4.10 & 1.86/3.89 
    & 1.72/3.06 & 1.53/2.95 & 1.45/2.86 \\
    \cmidrule(lr){1-9}
    
    \symbola AFormer-Mixed & $\rho^{S}$=$\rho^{M}$=$\rho^{L}$=0.5 & 6.67M 
    & 2.14/4.38 & 1.98/4.05 & 1.94/3.91 
    & 1.68/3.06 & 1.50/2.91	& 1.57/2.93\\
    
    \symbola AFormer-Mixed 
    & $\rho^{S}$=0.75, $\rho^{M}$=$\rho^{L}$=0.5 & 6.67M 
    & 2.10/4.38 & 1.98/4.02 & 2.05/4.14
    & 1.65/3.01 & 1.50/2.90 & 1.66/3.11 \\
    
    \symbolb AFormer-Tuning & 4Ts$\rightarrow$2Ts & 6.67M
    & 2.10/4.32 & 2.15/4.31 & 2.28/4.43 
    & 1.60/2.98 & 1.81/3.34	& 1.79/3.12 \\
    
    \symbolb AFormer-Tuning & 4Ts$\rightarrow$3Ts & 6.67M
    & 2.32/4.67 & \underline{1.93}/\underline{3.94} & 2.32/4.96
    & 1.98/3.33 & 1.48/2.92 & 2.12/3.97 \\
    
    \symbold AFormer-Joint 	& - & 6.67$\times$3M
    & \underline{1.98}/\underline{4.19} & 1.95/3.97 & 1.90/4.02	
    & 1.54/3.04 & 1.50/2.91	& 1.49/2.96 \\
    
    \symbolc AFormer-IT  & - & 6.67$\times$3M
    & 2.02/4.23 & \underline{1.93}/3.97 & \underline{1.86}/\underline{3.89}
    & \underline{1.52}/\underline{3.00} & \underline{1.47}/\underline{2.89}	 & \underline{1.45}/\underline{2.86} \\
    
     \cmidrule(lr){1-9}
    \textbf{AFormer-FLN} & $\rho^{S}$=$\rho^{M}$=$\rho^{L}$=1 & 6.68M
    & \textbf{1.92/3.91} & \textbf{1.88/3.89} & \textbf{1.83/3.78}	
    & \textbf{1.47/2.90} & \textbf{1.43/2.82} & \textbf{1.32/2.73}  \\ 
	\bottomrule
    \end{tabular}
     }
     \vspace{-1mm}
    \caption{Comparison with baseline models on nuScenes, evaluated using ADE$_{5}$/FDE$_{5}$ and ADE$_{10}$/FDE$_{10}$ metrics. The best results are highlighted in bold and the second-best results are underlined.}
    \vspace{-2mm}
\label{tab:nus}
\end{table*}

\begin{table*}[t]
	\centering
    \scalebox{0.75}{
    \begin{tabular}{lccccc}
    \toprule
     \multicolumn{1}{l}{\multirow{2}{*}{Method}} 
     & \multicolumn{1}{c}{\multirow{2}{*}{Note}}
     & \multicolumn{1}{c}{\multirow{2}{*}{\#Param}}
     & \multicolumn{3}{c}{ADE$_{6}$/FDE$_{6}$ $\downarrow$ \quad K = 6 Samples} \\
    \cmidrule(lr){4-6}
     & & & 10 Timesteps & 20 Timesteps & 30 Timesteps \\
    \midrule

    HiVT-64~\cite{zhou2022hivt} & - & 662K
    & 1.23/1.48	& 0.91/1.37	& 0.69/1.04\\
    \cmidrule(lr){1-6}
    \symbola HiVT-64-Mixed 
    & $\rho^{S}$=$\rho^{M}$=$\rho^{L}$=0.5 & 662K 
    & 1.14/1.57	& 0.86/1.22	& 0.80/1.12\\
    \symbola HiVT-64-Mixed 
    & $\rho^{S}$=0.75, $\rho^{M}$=$\rho^{L}$=0.5 & 662K 
    & 1.08/1.49	& 0.88/1.22	& 0.84/1.13\\
    
    \symbolb HiVT-64-Tuning & 30Ts$\rightarrow$10Ts & 662K 
    & \underline{0.90}/\underline{1.38} & 1.14/1.48 & 1.18/1.31 \\
    \symbolb HiVT-64-Tuning & 30Ts$\rightarrow$20Ts & 662K 
    & 1.33/1.61	& 0.87/1.20 & 1.12/1.29
     \\
    
    \symbold HiVT-64-Joint & - & 662$\times$3K
    & 0.98/1.45 & 0.89/1.24 & 0.74/1.09 \\
    
    \symbolc HiVT-64-IT  & - & 662$\times$3K
    & 0.92/1.43 & \underline{0.81}/\underline{1.17} & \underline{0.69}/\underline{1.04}\\

    \cmidrule(lr){1-6}
    \textbf{HiVT-64-FLN} & $\rho^{S}$=$\rho^{M}$=$\rho^{L}$=1 & 680K
    & \textbf{0.81/1.25} & \textbf{0.72/1.08} & \textbf{0.65/0.98} \\ 
	\bottomrule
    \end{tabular}
    }
    \vspace{-1mm}
    \caption{Comparison with baseline models on the Argoverse 1 validation set, evaluated using ADE$_{6}$/FDE$_{6}$ metrics. The best results are highlighted in bold and the second-best results are underlined.}
    \vspace{-2mm}
\label{tab:argo}
\end{table*}

\noindent
\textbf{Specialized Layer Normalization.} 
As outlined in \cref{sec:look}, normalization shift is another reason for the performance gap when observation length shifts. Denote the intermediate features for $X^{S}$, $X^{M}$, and $X^{L}$ as $f^{S}$, $f^{M}$, and $f^{L}$, respectively. We introduce the specialized layer normalization for each input sequence as follows:
\begin{equation}
    \label{eq:ln}
    y^{*}=\gamma^{*}\frac{f^{*}-\mu^{*}}{\sqrt{{\sigma^{*}}^{2}+\epsilon}} + \beta^{*},
\end{equation}
where $*\in\{S, M, L\}$. This specialized normalization allows for the independent learning of $\gamma$ and $\beta$, and the calculation of $\mu$ and $\sigma$ specific to each sequence during training. Moreover, this approach is efficient, as normalization typically involves a simple transformation with less than $1\%$ of total model parameters.
\begin{figure*}[t]
\centering
    \begin{subfigure}[b]{0.2\textwidth}
        \centering
        \includegraphics[width=\textwidth]{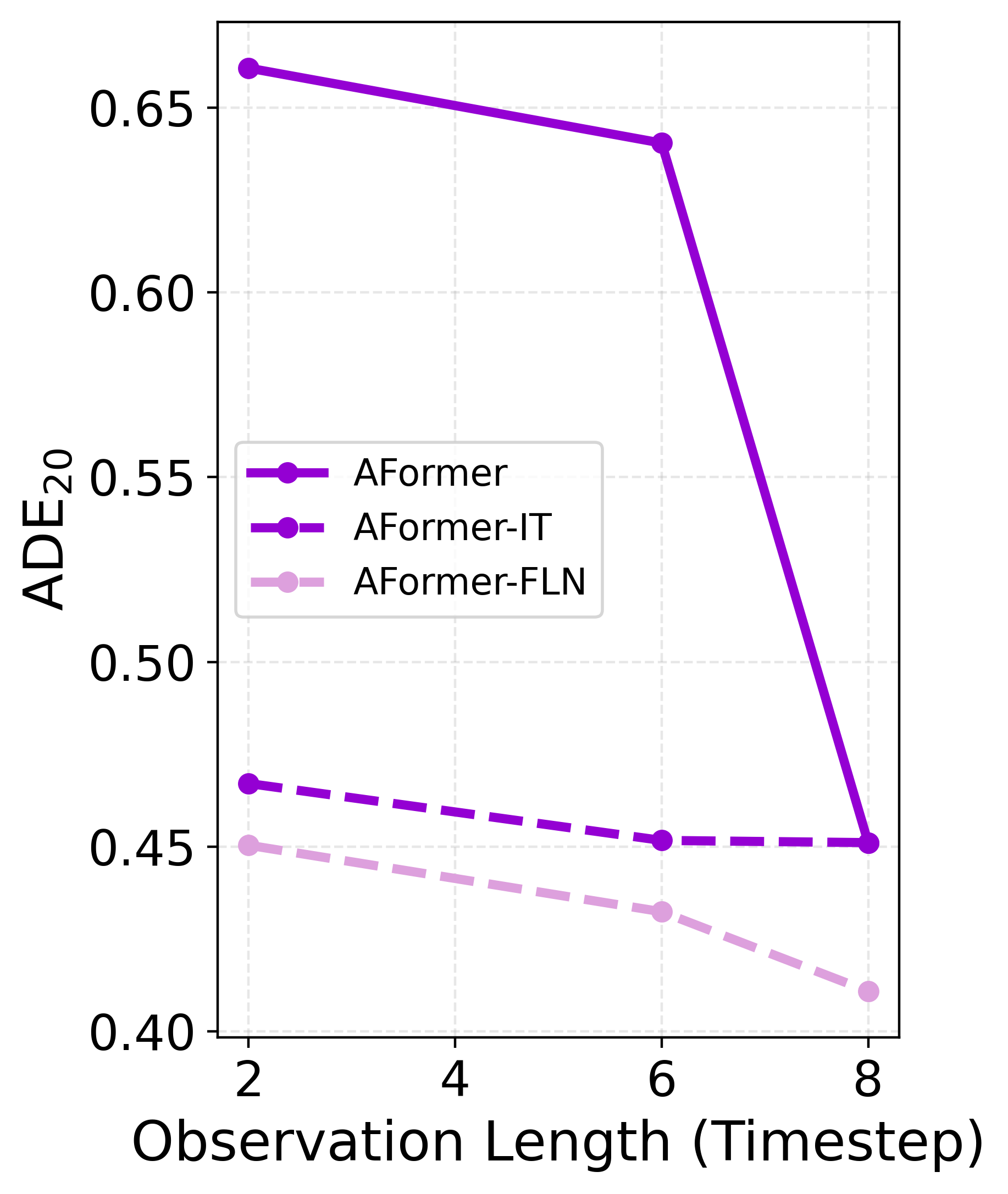}
        \caption{ADE$_{20}$ on Eth.}
    \end{subfigure}
    \hspace{-2.5mm}
    \begin{subfigure}[b]{0.2\textwidth}
        \centering
        \includegraphics[width=\textwidth]{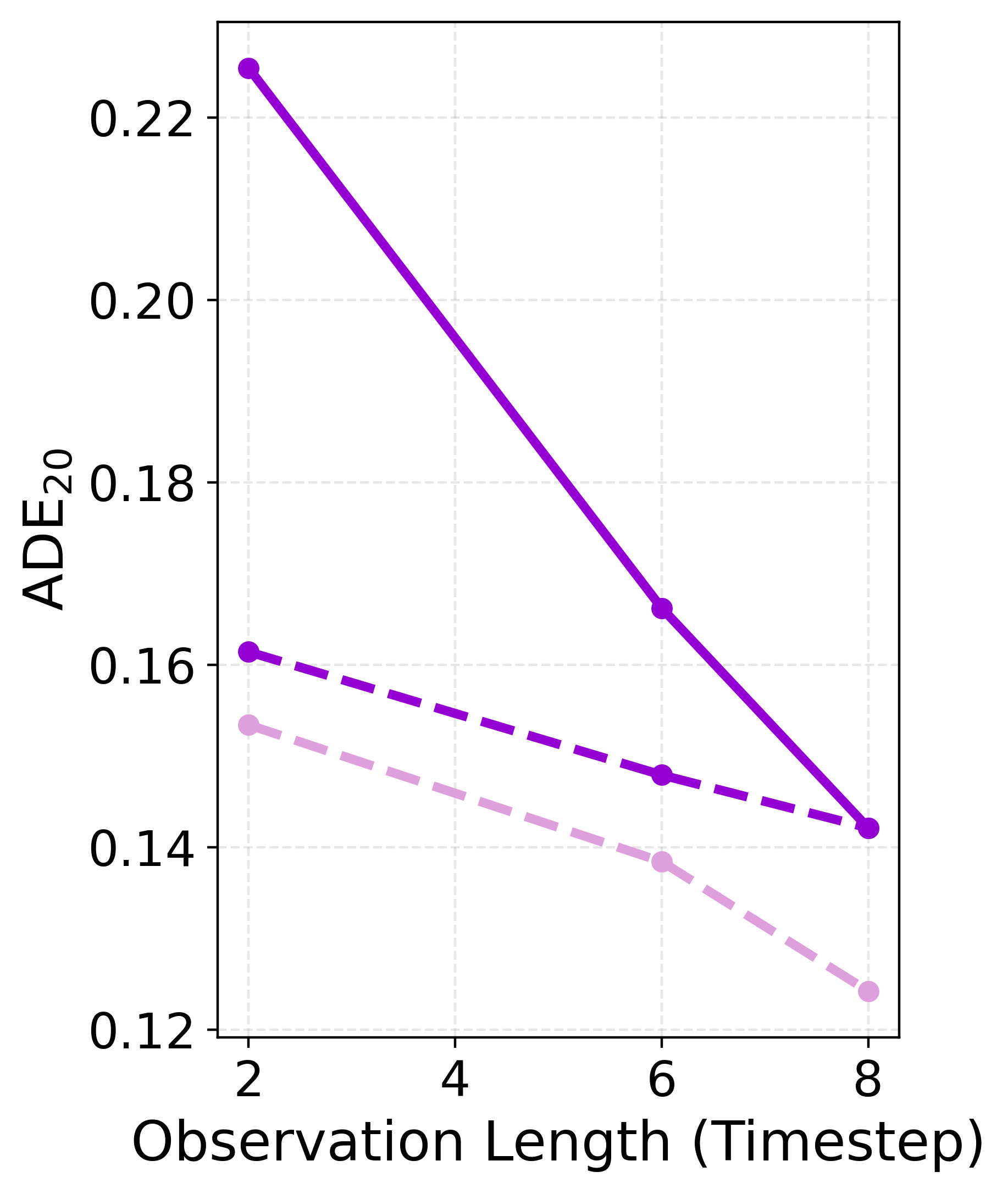}
        \caption{ADE$_{20}$ on Hotel.}
    \end{subfigure}
    \hspace{-2.5mm}
    \begin{subfigure}[b]{0.2\textwidth}
        \centering
        \includegraphics[width=\textwidth]{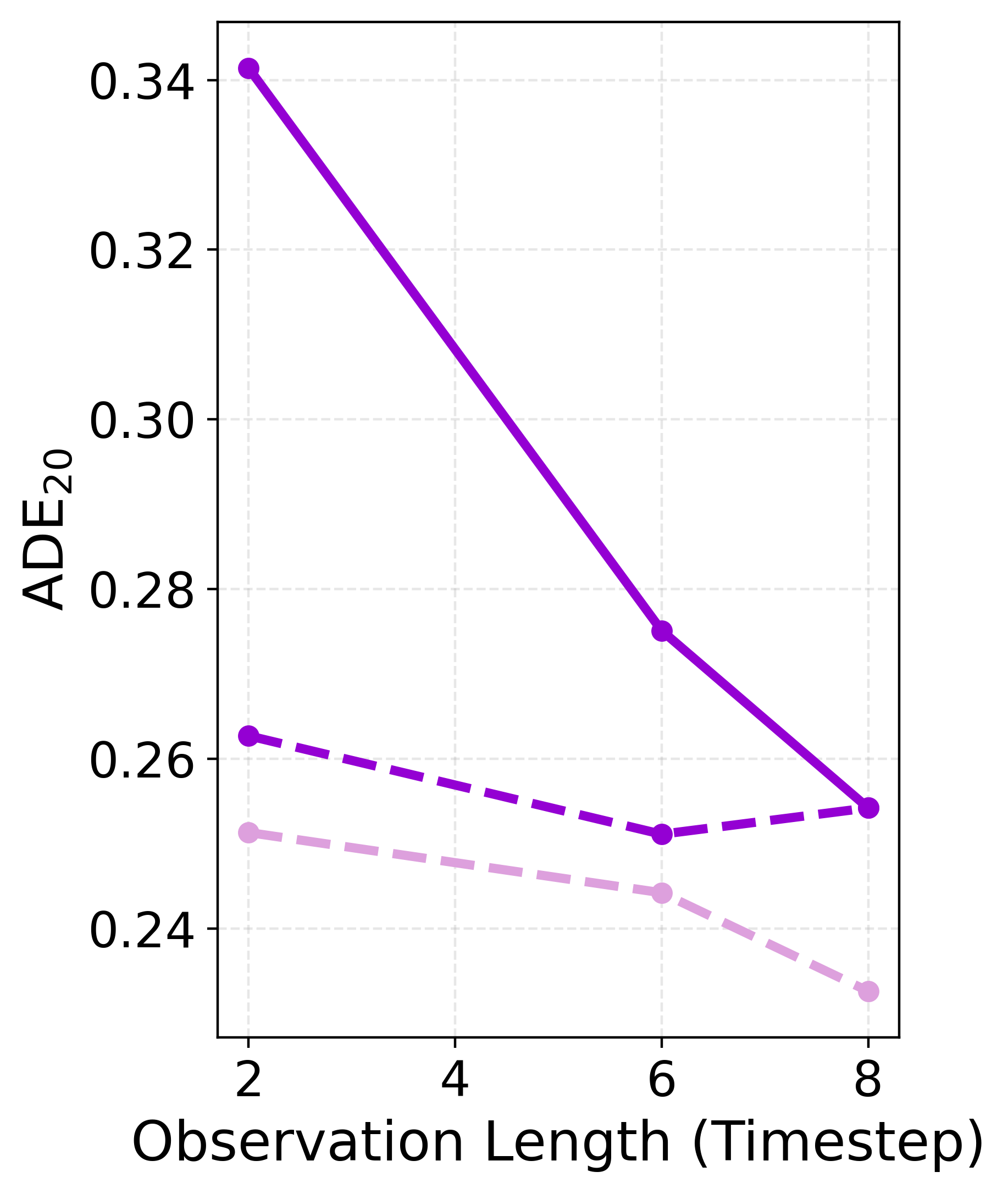}
        \caption{ADE$_{20}$ on Univ.}
    \end{subfigure}
    \hspace{-2.5mm}
    \begin{subfigure}[b]{0.2\textwidth}
        \centering
        \includegraphics[width=\textwidth]{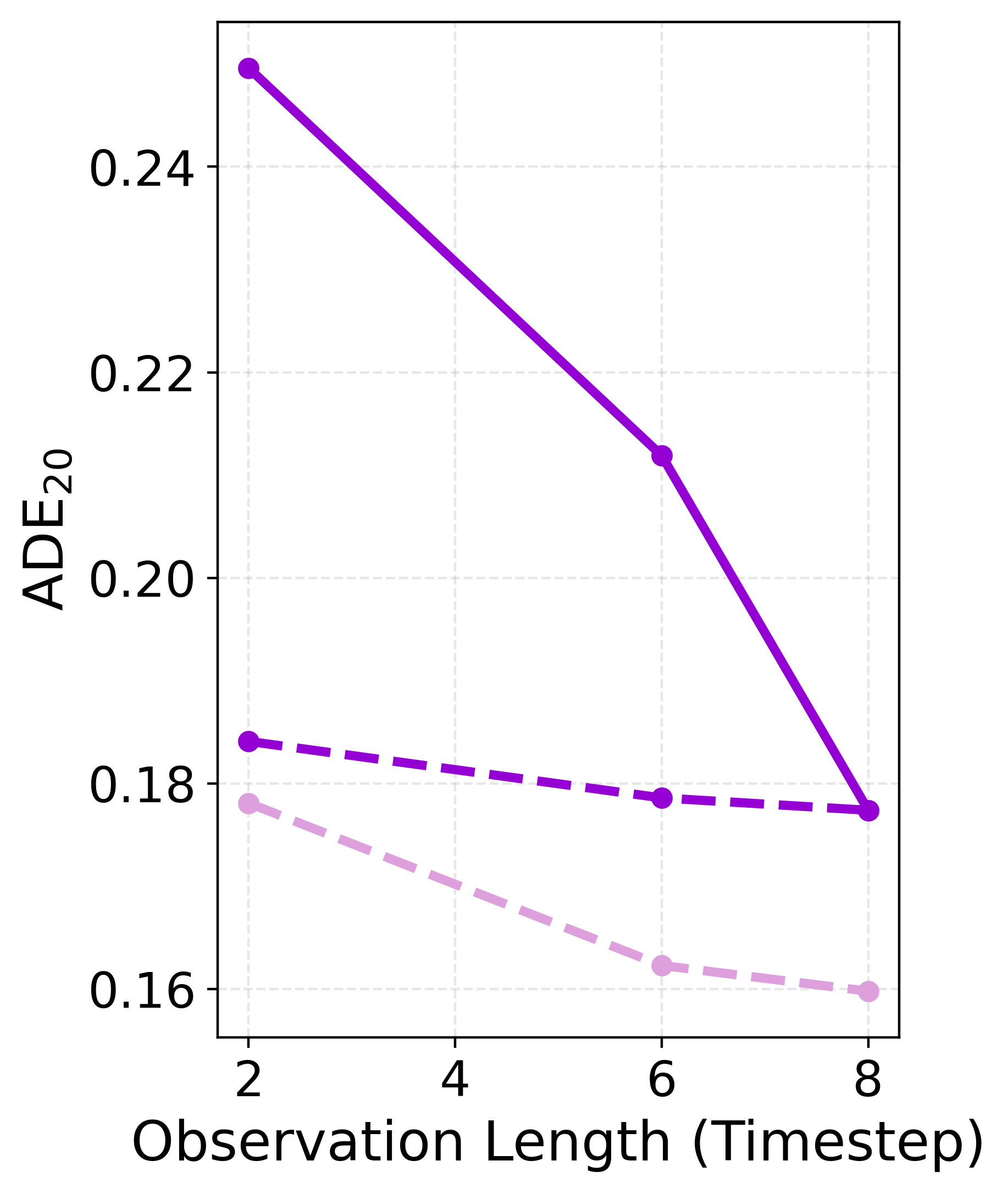}
        \caption{ADE$_{20}$ on Zara1.}
    \end{subfigure}
    \hspace{-2.5mm}
    \begin{subfigure}[b]{0.2\textwidth}
        \centering
        \includegraphics[width=\textwidth]{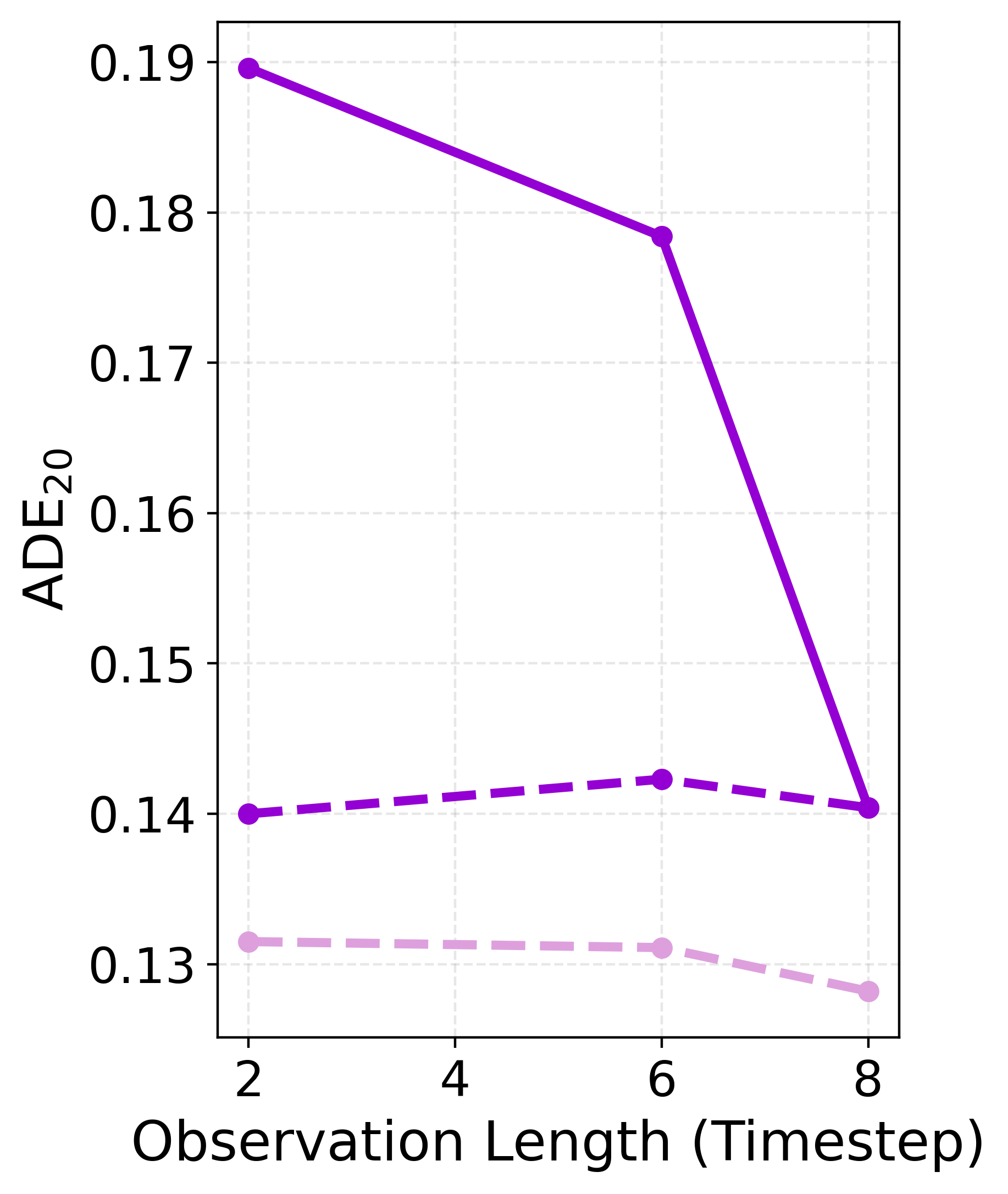}
        \caption{ADE$_{20}$ on Zara2.}
    \end{subfigure}
    \vspace{-1mm}
    \caption{Performance on five ETH/UCY datasets using the AgentFormer model, measured with ADE$_{20}$. These results are compared with those of the baseline model and Isolated Training (IT), showcasing notable improvements achieved by our FLN.}
    \vspace{-2mm}
    \label{fig:ethucy_ade}
\end{figure*}

\begin{figure*}[t]
\centering
    \begin{subfigure}[b]{0.2\textwidth}
        \centering
        \includegraphics[width=\textwidth]{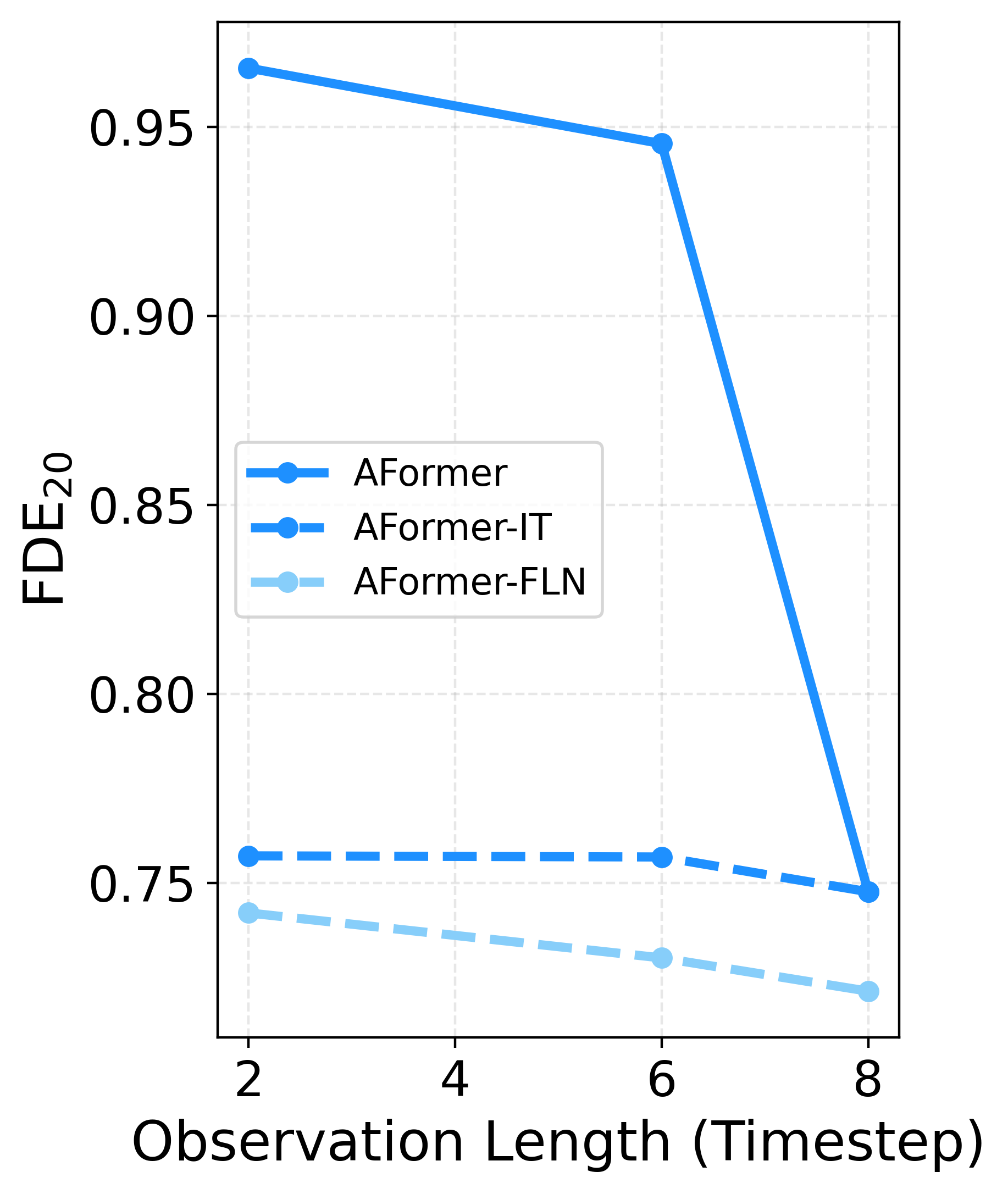}
        \caption{FDE$_{20}$ on Eth.}
    \end{subfigure}
    \hspace{-2.5mm}
    \begin{subfigure}[b]{0.2\textwidth}
        \centering
        \includegraphics[width=\textwidth]{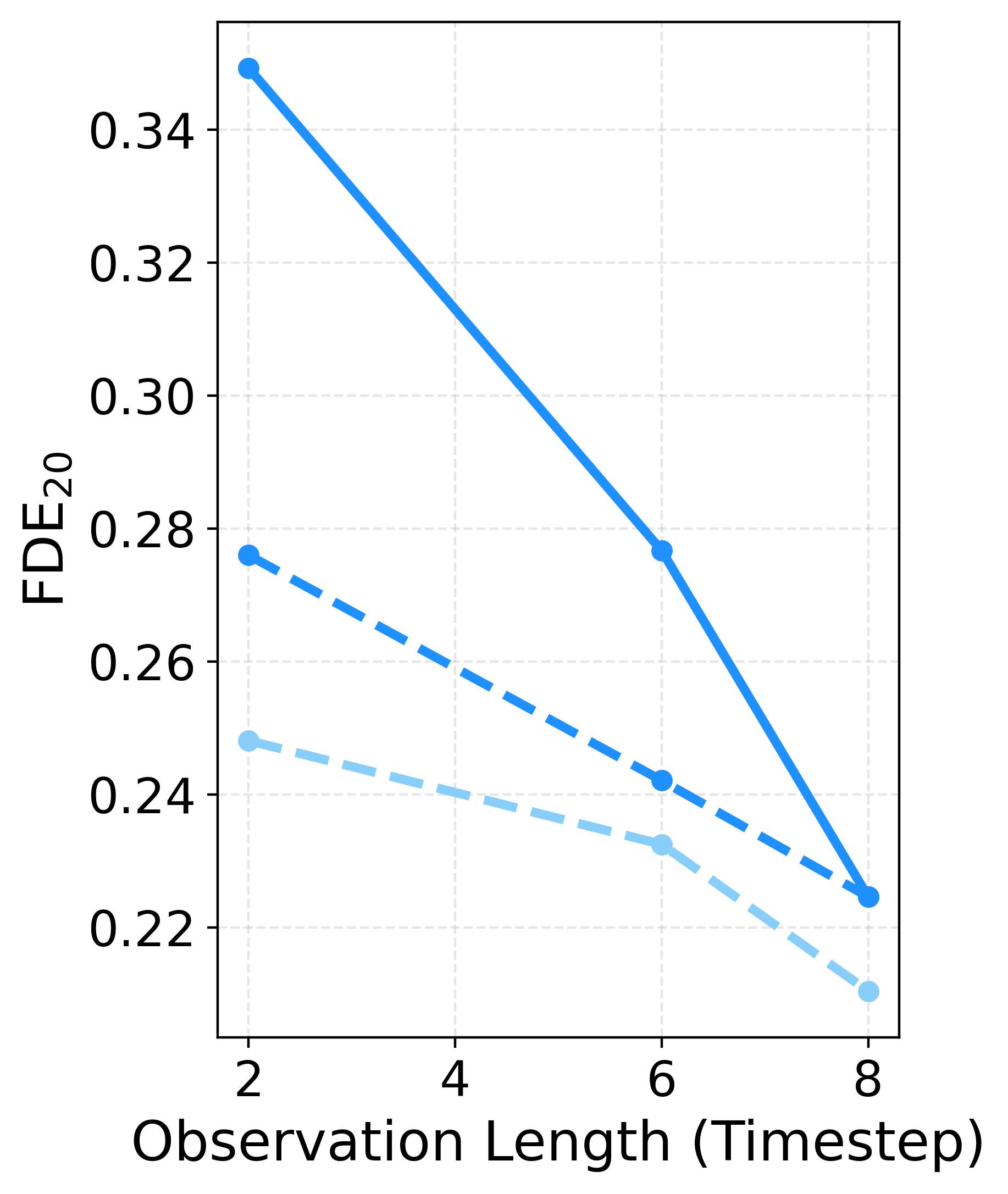}
        \caption{FDE$_{20}$ on Hotel.}
    \end{subfigure}
    \hspace{-2.5mm}
    \begin{subfigure}[b]{0.2\textwidth}
        \centering
        \includegraphics[width=\textwidth]{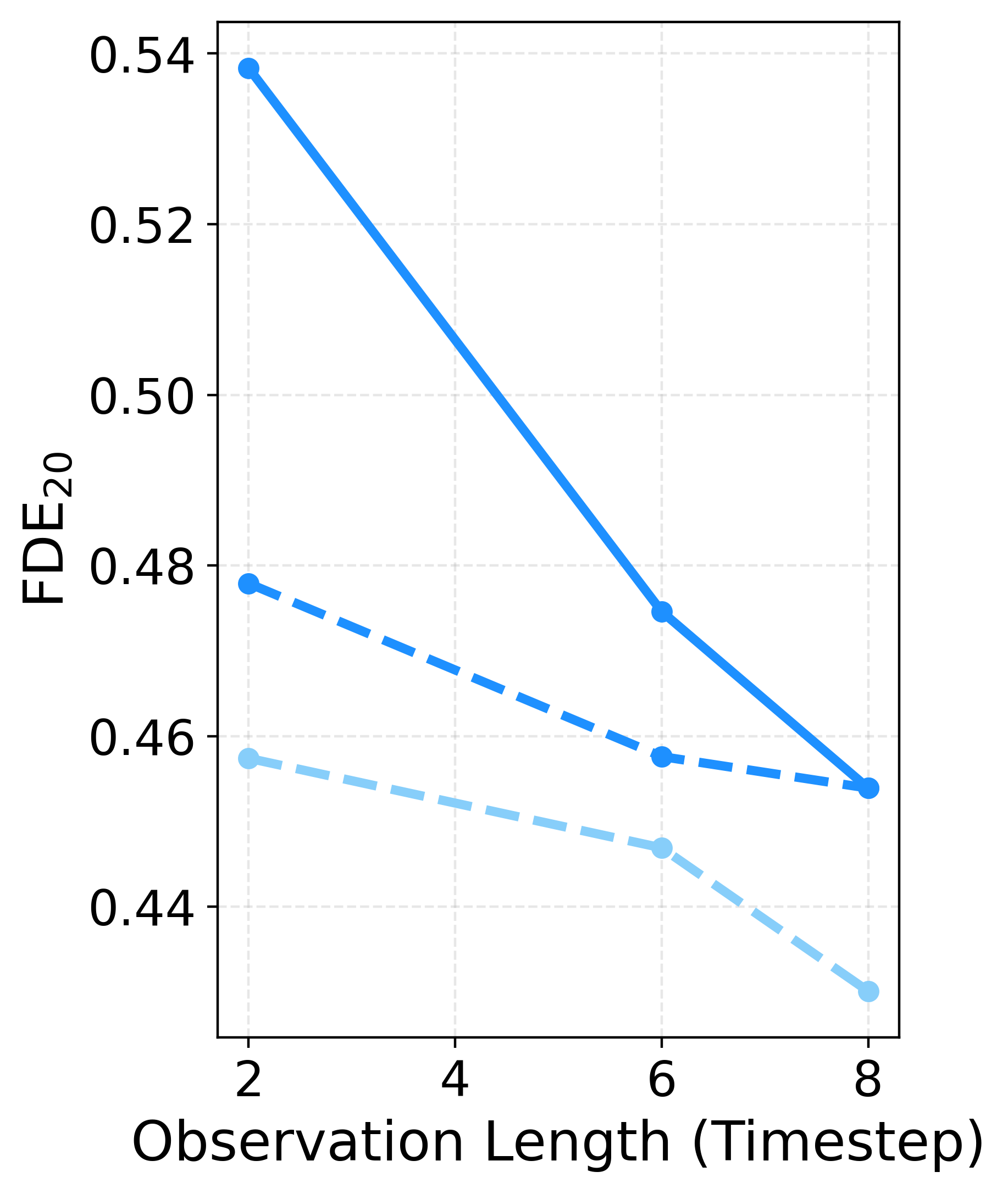}
        \caption{FDE$_{20}$ on Univ.}
    \end{subfigure}
    \hspace{-2.5mm}
    \begin{subfigure}[b]{0.2\textwidth}
        \centering
        \includegraphics[width=\textwidth]{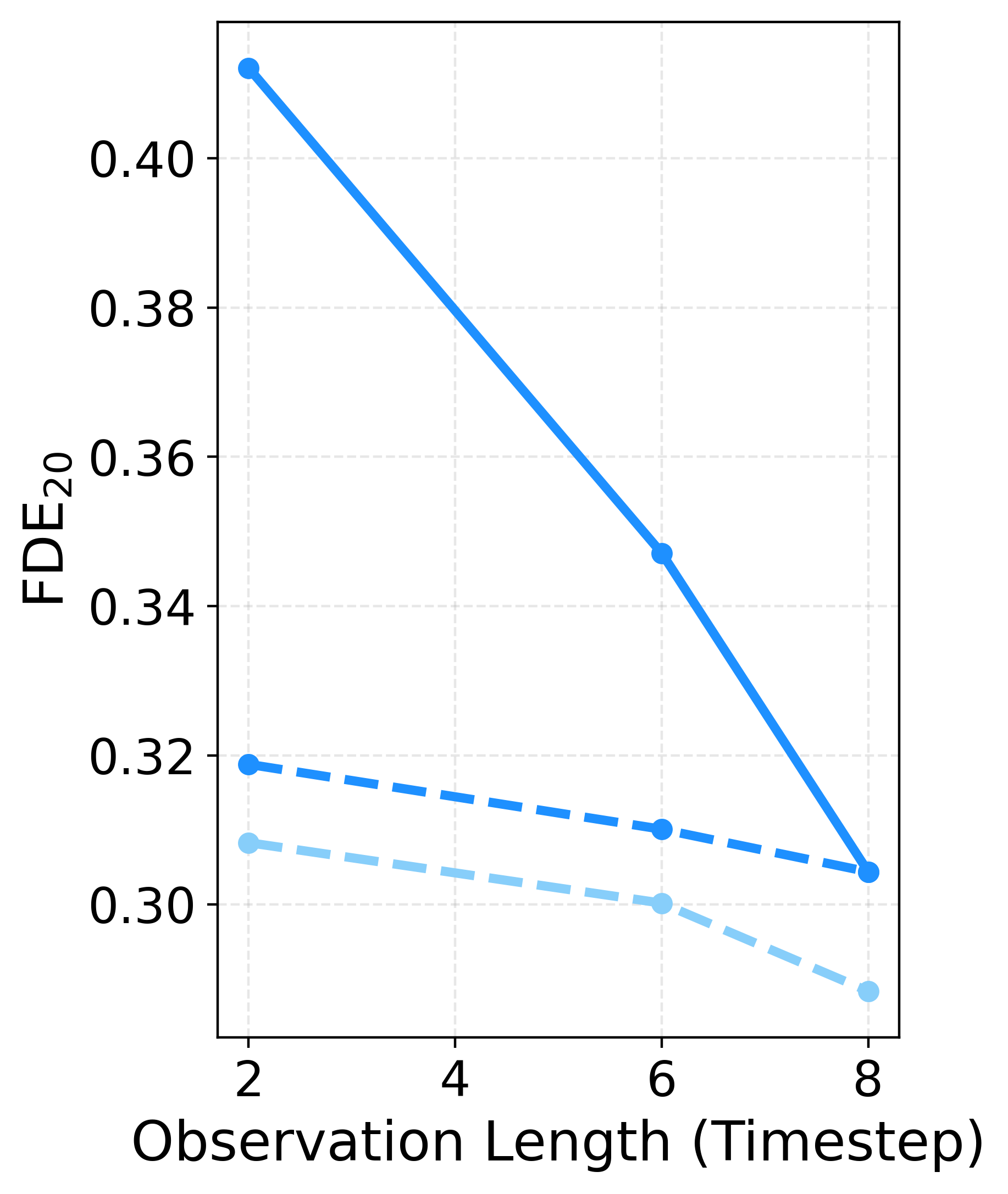}
        \caption{FDE$_{20}$ on Zara1.}
    \end{subfigure}
    \hspace{-2.5mm}
    \begin{subfigure}[b]{0.2\textwidth}
        \centering
        \includegraphics[width=\textwidth]{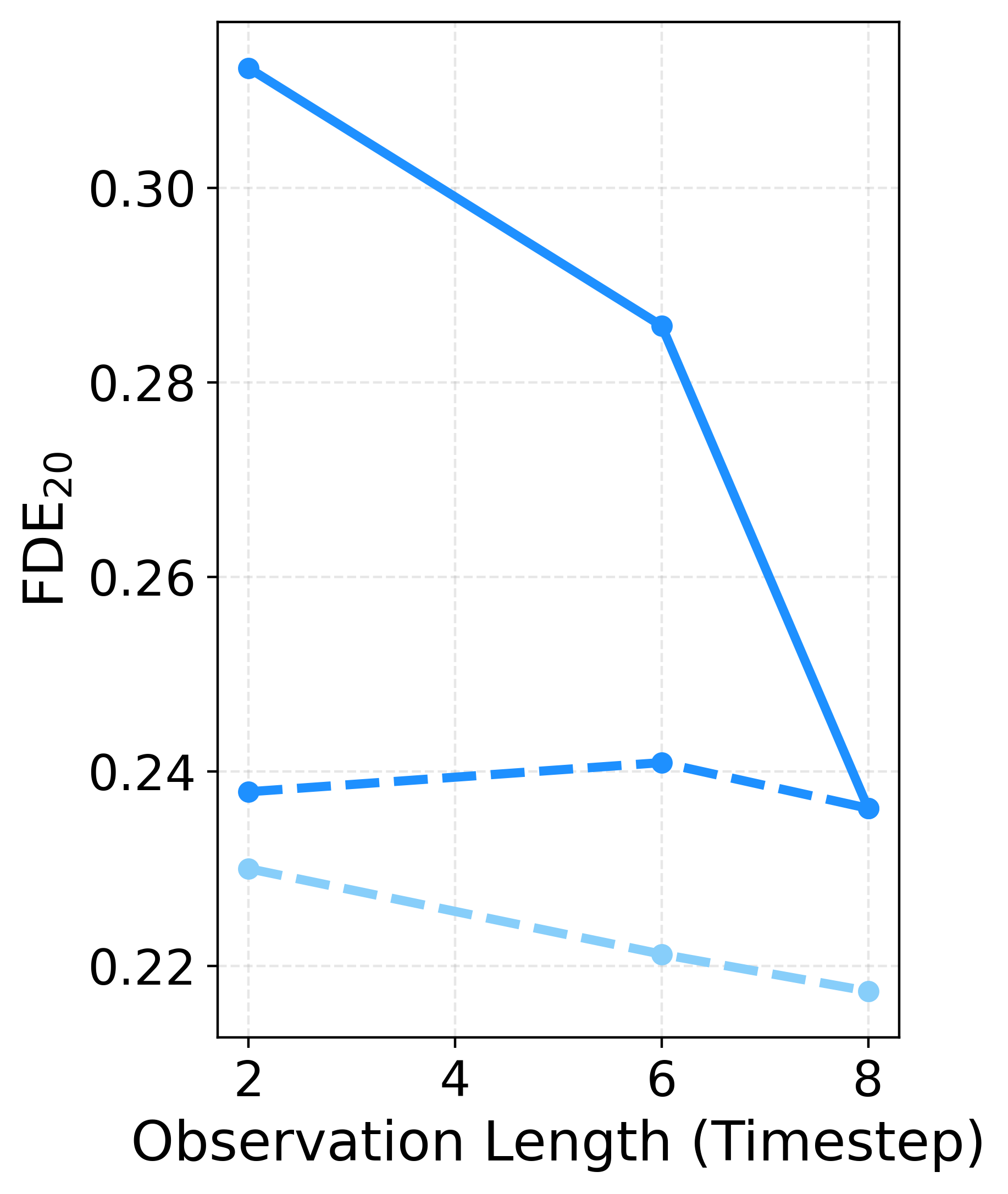}
        \caption{FDE$_{20}$ on Zara2.}
    \end{subfigure}
    \vspace{-1mm}
    \caption{Performance on five ETH/UCY datasets using the AgentFormer model, measured with FDE$_{20}$. These results are compared with those of the baseline model and Isolated Training (IT), showcasing notable improvements achieved by our FLN.}
    \vspace{-2mm}
    \label{fig:ethucy_fde}
\end{figure*}

\section{Experiments}
\label{sec:exp}
\subsection{Settings}
\noindent\textbf{Baselines.}
Our proposed FLN is a general framework that is compatible with those Transformer-based methods. We have chosen two widely recognized open-source methods, AgentFormer~\cite{yuan2021agentformer} and HiVT~\cite{zhou2022hivt}, to integrate with our FLN. Furthermore, for a more comprehensive comparison, we include four additional baseline models:
(1) \textbf{Mixed Sampling}: In each training iteration, we assign three probabilities $\rho^{S}$, $\rho^{M}$, and $\rho^{L}$ to take trajectory data with observation lengths $H^{S}$, $H^{M}$, and $H^{L}$ for training. (2) \textbf{Fine-tuning}: The model is trained using the observation length $H^{L}$, and then fine-tuned using another length ($H^{S}$ or $H^{M}$) until convergence.
(3) \textbf{Joint}: We expand the training dataset by including trajectory samples from all three observation lengths ($H^{S}$, $H^{M}$, and $H^{L}$) and then train the original model without any structural changes.
(4) \textbf{Isolated Training}: The model is trained exclusively with trajectory data of a single observation length, $H^{S}$, $H^{M}$, or $H^{L}$.

\noindent\textbf{Datasets.}
We use three following datasets:
(1) The ETH/UCY dataset~\cite{pellegrini2009You, lerner2007crowds} is a primary benchmark for pedestrian trajectory prediction, including five datasets, Eth, Hotel, Univ, Zara1, and Zara2, with densely interactive trajectories sampled at 2.5Hz.
(2) The nuScenes dataset~\cite{zhan2019interaction} is a large autonomous driving dataset featuring 1000 scenes, each annotated at 2Hz, and includes HD maps with 11 semantic classes.
(3) Argoverse 1~\cite{chang2019argoverse} contains 323,557 real-world driving sequences sampled at 10Hz, complemented with HD maps for trajectory prediction.

\noindent\textbf{Evaluation Protocol.}
Across the three datasets, we use the metrics Average Displacement Error (ADE$_{K}$) and Final Displacement Error (FDE$_{K}$) for comparison, where $K$ indicates the number of trajectories to be predicted. Each dataset follows its own evaluation protocol:
(1) For ETH/UCY, the leave-one-out setting is standard, with the task to predict 12 future time steps from 8 observed steps. Here, $K$ is commonly set to 20.
(2) Within the nuScenes dataset, as used in AgentFormer, only vehicle data is considered, predicting 12 future steps from 4 observed steps. $K$ is usually set to 5 and 10.
(3) In Argoverse 1, sequences are segmented into 5-second intervals, with predictions made for 30 future steps (3 seconds) based on 20 observed steps (2 seconds) involving multiple agents. The validation set is used for our evaluation purpose, with $K$ being 6.

\noindent
\textbf{Implementation Details.}
For each dataset, we define three different observation lengths (in timestep) $\mathbf{H}=\{H^{S}, H^{M}, H^{L}\}$ for training, where $H^{L}$ is set as the default length in the standard evaluation protocol, considering the potential unavailability of data beyond this length. Specifically, we use $\mathbf{H}=\{2, 6, 8\}$ for ETH/UCY, $\mathbf{H}=\{2, 3, 4\}$ for nuScenes, and $\mathbf{H}=\{10, 20, 30\}$ for Argoverse 1. The prediction length $T$ remains as defined in the standard settings. The prototype models are trained at observation length $H^{L}$ and evaluated at $H^{S}$ and $H^{M}$. Isolated Training (IT) refers to training and evaluating the model separately at observation length $H^{S}$, $H^{M}$, and $H^{L}$. More details and experiments are provided in the Supplementary Material.

\subsection{Main Results}
\noindent
\textbf{Comparison with Baselines.}
\cref{tab:nus} presents the results using the AgentFormer model on the nuScenes dataset, where our FLN framework outperforms all baseline models, showcasing its robustness in adapting to different observation lengths. It is noticeable that Mixed Sampling helps mitigate the Observation Length Shift issue, as indicated by the improvement at observation lengths of 2 and 3 timesteps, compared to the results achieved with the standard training protocol. However, this improvement comes at the cost of reduced accuracy at the 4-timestep observation length. Increasing the probability for length $H^{S}$ improves the performance at that length, but conversely, the performance at $H^{L}$ becomes worse. This pattern suggests a trade-off: improved results at shorter lengths result in diminished performance at longer lengths. One interesting observation is that the performance at $H^{M}$ remains relatively stable, possibly because the quantity of sequence data for $H^{M}$ and $H^{L}$ remains unchanged, thus maintaining a consistent relationship between these two data clusters. Fine-tuning enables the baseline method to achieve comparable performance with IT, neglecting information from other lengths. As for the joint baseline, its performance is similar to IT, and even better than IT at the 2-timestep length when $K=5$. However, a significant issue with both Joint and IT baselines is the increased complexity of the models due to repeated training sessions. In contrast, our FLN requires only a single training session with minimal additional computational overhead, yet it allows for evaluation at various lengths and obtains superior results.

\noindent
\textbf{Performance across Datasets.}
Additionally, the evaluation on the ETH/UCY benchmark, as illustrated in \cref{fig:ethucy_ade} and \cref{fig:ethucy_fde}, reveals that our proposed FLN outperforms IT at different observation lengths. We also extend our FLN to the HiVT method, which utilizes learnable embeddings for positional encoding, and evaluate it on the Argoverse 1 validation set. The results, presented in \cref{tab:argo}, show that FLN consistently achieves superior performance across three different observation lengths, affirming its robustness in adapting to various observation lengths.

\subsection{Generality Study}
\begin{figure}[t]
  \centering
   \includegraphics[width=0.85\linewidth]{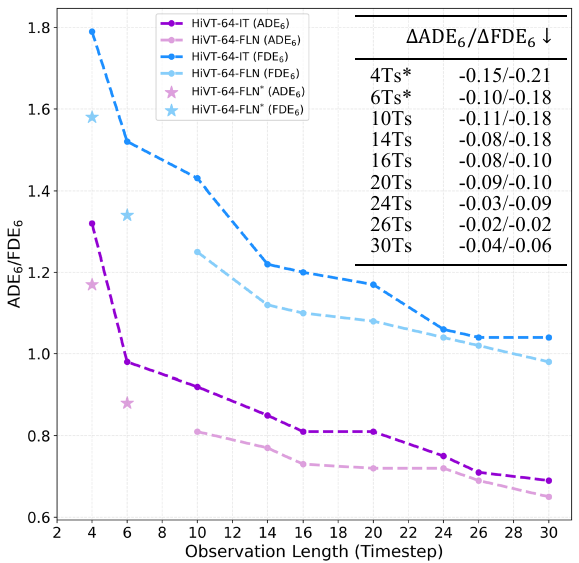}
   \vspace{-1mm}
   \caption{Performance of FLN and IT at different observation lengths on the Argoverse 1 validation set. The improvements are listed in the table. Results from observation lengths outside the range of 10 to 30 are denoted with a $*$.}
   \vspace{-4mm}
   \label{fig:argo}
\end{figure}
We have demonstrated that FLN surpasses IT at the training observation lengths, while not evaluating the model at lengths outside the training range. Inspired by our findings in \cref{sec:look}, where the performance gap narrows with observation length closer to the standard input length, we introduce an efficient inference approach enabling FLN to be evaluated at any length. When encountering an unseen length $H'$, we first calculate its difference from $H^{S}$, $H^{M}$, and $H^{L}$, and then activate the sub-network with the least difference in length. In cases where two sub-networks have the same length difference, we default to the one with the longer input length. This approach allows for the evaluation of FLN across a spectrum of unseen observation lengths.

Since we set the longest length $H^{L}$ as the standard observation length, it is challenging to generate trajectory sequences with observation lengths longer than $H^{L}$ due to their pre-segmentation. The results on the Argoverse 1 validation set with observation length shorter than 30 timesteps are illustrated in \cref{fig:argo}. These results clearly demonstrate that FLN consistently outperforms IT at all observation lengths, even those beyond the 10-30 range. This validates the strong generality of our FLN in handling observation lengths that are not included during training.

\subsection{Ablation Study}
\noindent
\textbf{Length Combination.} 
We implement different combinations of observation lengths on Argoverse 1, with the results presented in \cref{tab:com}. It is observed that FLN-2 surpasses IT in terms of ADE$_{6}$ and FDE$_{6}$ at the length of 10 and 16 timesteps, and achieves comparable results at the length of 30. However, its performance at 20-timestep length is worse than IT, likely due to the absence of this sequence length in the training data. Both FLN-3 and FLN-4 outperform IT at all observation lengths. This improvement can be attributed to the existence of trajectory data with intermediate observation lengths, which helps mitigate the Observation Length Shift within this range. Although FLN-4 shows better results than FLN-3 at lengths of 10, 16, and 20 timesteps, it also costs additional time and resources for training.
\begin{table}[t]
    \centering
    \scalebox{0.68}{
    \begin{tabular}{lccccc}
    \toprule
    \multicolumn{1}{l}{\multirow{2}{*}{Method}} 
    & \multicolumn{1}{c}{\multirow{2}{*}{Combination}}
    & \multicolumn{4}{c}{ADE$_{6}$/FDE$_{6}$ $\downarrow$ \quad K = 6 Samples} \\
    \cmidrule(lr){3-6}
    & & 10Ts & 16Ts & 20Ts & 30Ts\\
    \midrule
         HiVT-64-IT~\cite{zhou2022hivt} & -
         & 0.92/1.43
         & 0.81/1.20
         & 0.81/1.17
         & 0.69/1.04 \\
    \hdashline
         HiVT-64-FLN-2 
         & 10/30
         & 0.88/1.37	
         & 0.80/1.20	
         & 0.84/1.21	
         & 0.69/1.02   \\

         HiVT-64-FLN-3
         & 10/20/30
         & 0.81/1.25
         & 0.73/1.10
         & \textbf{0.72}/1.08
         & \textbf{0.65}/\textbf{0.98} \\

         HiVT-64-FLN-4 
         & 10/16/20/30
         & \textbf{0.79}/\textbf{1.19} 
         & \textbf{0.70}/\textbf{1.02}
         & \textbf{0.72}/\textbf{1.06}
         & 0.68/0.99 \\
        \bottomrule
    \end{tabular}
    }
    \vspace{-1mm}
    \caption{Performance of FLN on the Argoverse 1 validation set using different combinations of observation lengths. The best results are highlighted in bold.}
    \vspace{-1mm}
    \label{tab:com}
\end{table}

\begin{table}[t]
    \centering
    \scalebox{0.75}{
    \begin{tabular}{lcccc}
    \toprule
    \multicolumn{1}{l}{\multirow{2}{*}{Method}} 
    & \multicolumn{1}{c}{\multirow{2}{*}{Note}}
    & \multicolumn{3}{c}{ADE$_{5}$/FDE$_{5}$ $\downarrow$ \quad K = 5 Samples} \\
    \cmidrule(lr){3-5}
    & & 2Ts & 3Ts & 4Ts \\
    \midrule
         AFormer-IT~\cite{yuan2021agentformer} & -
         & 2.02/4.23 
         & 1.93/3.97	
         & 1.86/3.89\\
         \hdashline
         AFormer-FLN
         & w/o WS
         & 1.94/3.95		
         & 1.89/3.92
         & 1.92/3.94 \\
         
         AFormer-FLN
         & w/o TD
         & 2.23/4.39		
         & 1.98/4.01
         & 1.85/3.81 \\

         AFormer-FLN
         & w/o IPE
         & 2.13/4.33		
         & 2.02/4.07
         & 1.95/3.98 \\
         AFormer-FLN 
         & w/o SLN
         & 2.12/4.30	
         & 1.98/4.02	
         & 1.93/3.95 \\

        \cmidrule(lr){1-5}
         AFormer-FLN & -
         & 1.92/3.91	
         & 1.88/3.89	
         & 1.83/3.78 \\
        \bottomrule
    \end{tabular}
    }
    \vspace{-1mm}
    \caption{Ablation study of FLN on nuScenes. WS, TD, IPE, and SLN denote Sub-Network Weight Sharing, Temporal Distillation, Independent Positional Encoding, and Specialized Layer Normalization, respectively. The best results are highlighted in bold.}
    \vspace{-4mm}
    \label{tab:des}
\end{table}

\noindent
\textbf{Model Design.}
We conducted a detailed analysis of each component within our FLN on the nuScenes dataset, with results presented in \cref{tab:des}. We first remove Weight Sharing (w/o WS), creating unique sub-networks for each length. The results are better than IT at observation lengths of 2 and 3 timesteps, but worse at length of 4. This suggests that shared weights enable FLN to more effectively capture temporal-invariant features across diverse input lengths. We excluded Temporal Distillation (w/o TD) from FLN, meaning not utilizing the KL loss for optimization. This leads to the performance worse than IT at observation lengths of 2 and 3 timesteps, suggesting the necessity of our TD design. We also tested shared Positional Encoding (w/o IPE) and shared Layer Normalization (w/o SLN). Both resulted in a performance decrease, validating the significance of our Independent Positional Encoding (IPE) and Specialized Layer Normalization (SLN) designs.

\section{Conclusion}
\label{sec:con}
In this paper, we tackle the critical challenge of Observation Length Shift in trajectory prediction by introducing the FlexiLength Network (FLN). This novel framework, incorporating FlexiLength Calibration (FLC) and FlexiLength Adaptation (FLA), offers a general solution for handling varying observation lengths and requires only one-time training. Our thorough experiments on datasets such as ETH/UCY, nuScenes, and Argoverse 1 demonstrate that FLN not only improves prediction accuracy and robustness over a range of observation lengths but also consistently outperforms Isolated Training (IT).

\textbf{Limitation and Future Work.} One limitation is that FLN will increase training time due to handling several input sequences per training iteration. Going forward, we will focus on enhancing the training efficiency of FLN.


\clearpage
\setcounter{page}{1}
\maketitlesupplementary

\section{Implementation Details}
We employ the official codes of the AgentFormer model~\cite{yuan2021agentformer}, as found in~\footnote{\url{https://github.com/Khrylx/AgentFormer}}, and the HiVT~\cite{zhou2022hivt} model, as found in~\footnote{\url{https://github.com/ZikangZhou/HiVT}}, to  evaluate our proposed framework. We utilize the provided pre-trained models to assess the performance across different observation lengths. Specifically, for the HiVT model, we opt for its smaller variant, HiVT-64, and use the Argoverse 1 validation set. All models are trained on NVIDIA Tesla V100 GPUs, adhering to the same hyperparameters as specified in their respective official implementations.

\noindent
\textbf{Training Loss.}
In our work, we introduce the FlexiLength Network (FLN) framework, designed for easy integration with Transformer-based trajectory prediction models. We demonstrate its application by evaluating it using two models: AgentFormer and HiVT. The AgentFormer model is a two-stage generative model. In both stages, we utilize the output $Y^{L}\sim \mathcal{D}(\psi^{L})$, following their original loss function. Additionally, we incorporate our temporal distillation loss $\mathcal{L}_{kl}$, setting the balance hyperparameter $\lambda$ to 1. For the HiVT model, which is trained using a combined regression and classification loss function, we also apply the output $Y^{L}\sim \mathcal{D}(\psi^{L})$ in line with its original loss function. Our framework further extends this with our temporal distillation loss $\mathcal{L}_{kl}$, and sets the balance hyperparameter $\lambda$ to 1.

\section{Specialized Layer Normalization Study}
In our analysis, we classify the components of typical Transformer-based models for trajectory prediction into several key parts: a Spatial Encoder for extracting spatial features, a Positional Encoder (PE) for embedding positional information, a Transformer Encoder for temporal dependency modeling, and a Trajectory Decoder for generating predicted trajectories. While various designs differ among these components, including Layer Normalization (LN) layers beyond the Transformer Encoder, we investigate these LN layers in different model parts. Our experiments show that the LN shift in the Transformer Encoder is the main cause of the performance drop.

\begin{figure}[t]
\centering
    \begin{subfigure}[b]{0.23\textwidth}
        \centering
        \includegraphics[width=\textwidth]{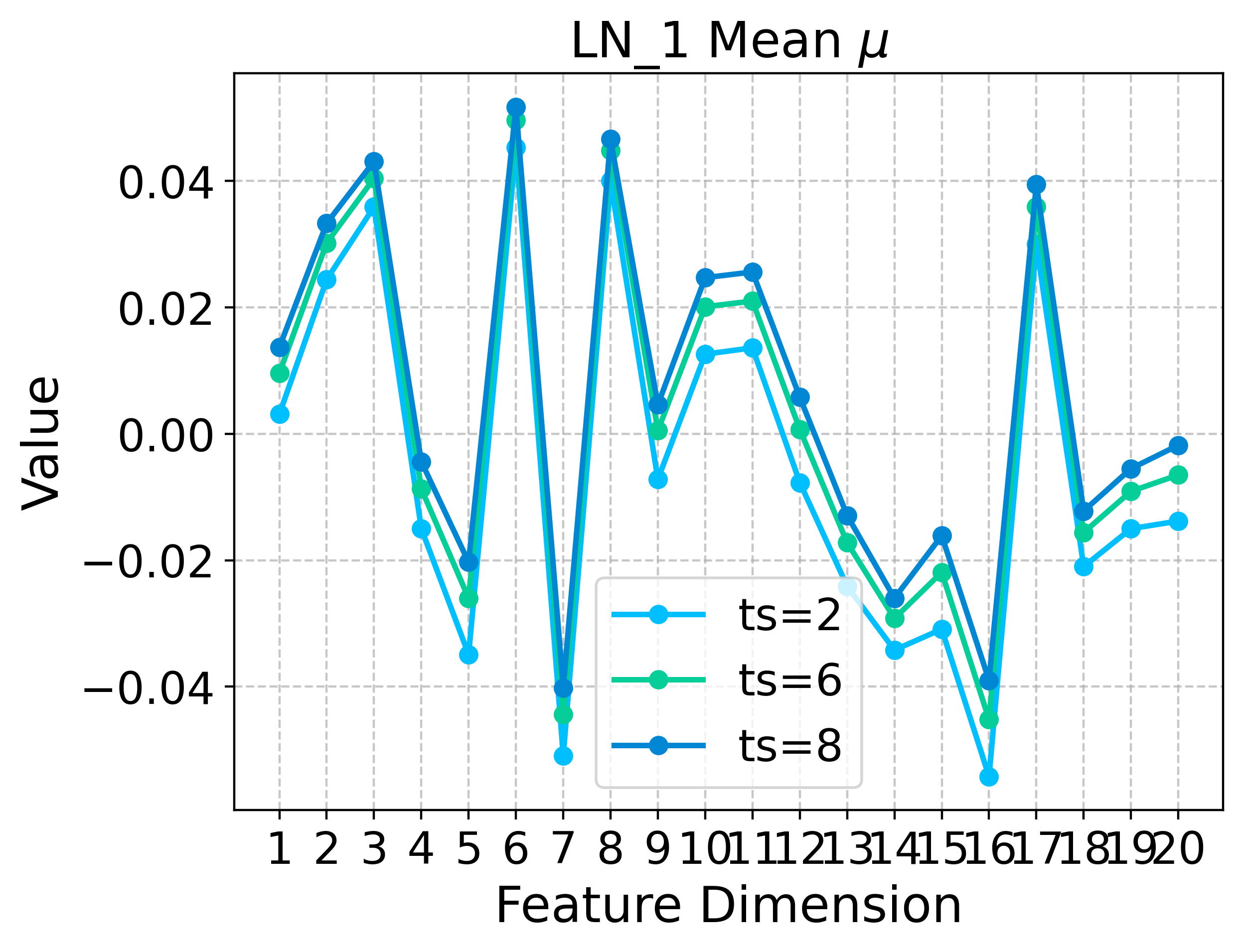}
    \end{subfigure}
    \begin{subfigure}[b]{0.23\textwidth}
        \centering
        \includegraphics[width=\textwidth]{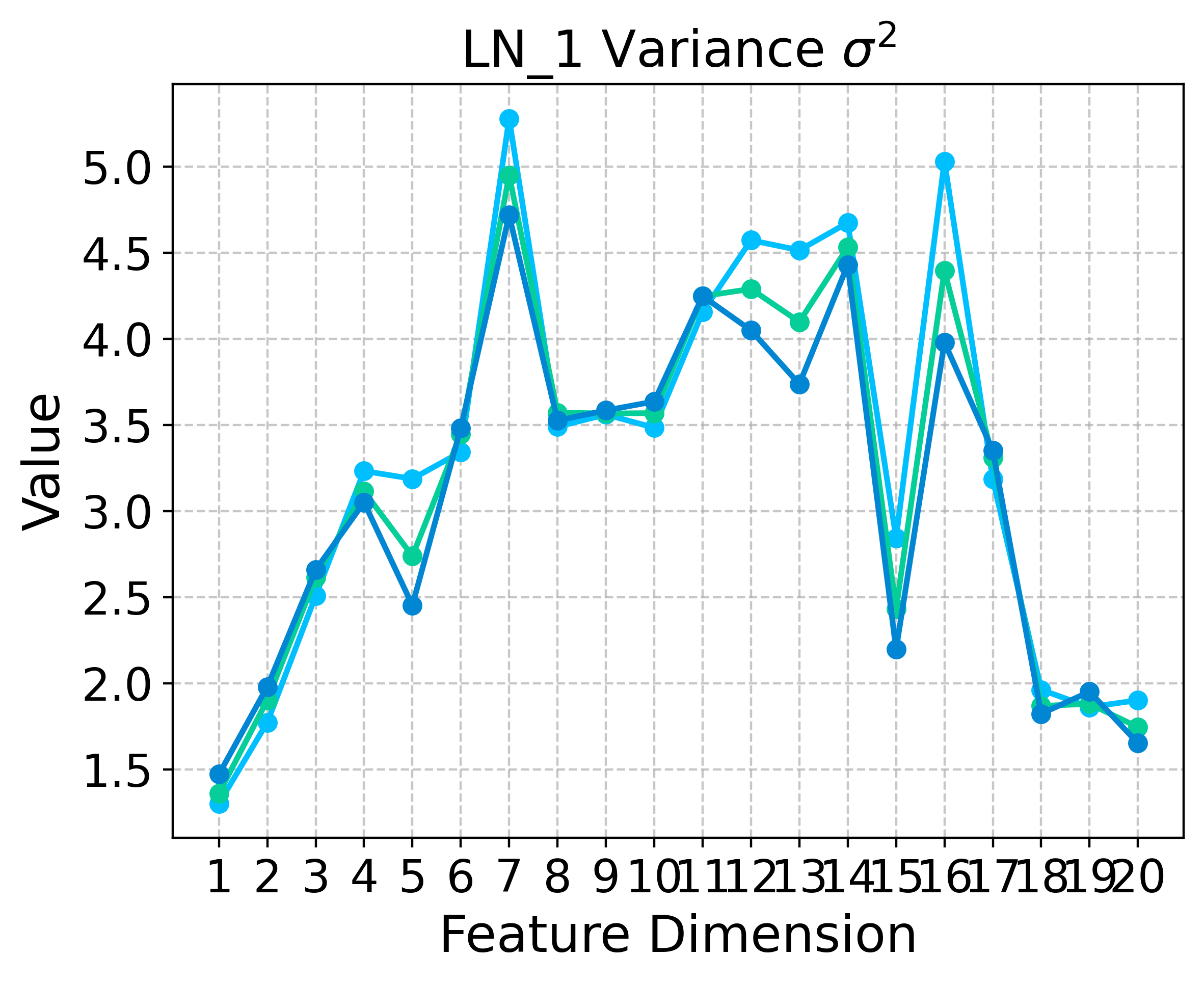}
    \end{subfigure}
    \caption{Layer Normalization statistics for the first LN layer of the Trajectory Decoder in the AgentFormer model,  trained (Isolated Training) on the Eth dataset with observation lengths of 2, 6, and 8 timesteps separately}
    \label{fig:ln_decoder}
\end{figure}

Regarding the AgentFormer model, it incorporates two LN layers in its Transformer Encoder and three in the Trajectory Decoder. We train the AgentFormer model separately for observation lengths of 2, 6, and 8 timesteps (Isolated Training) on the Eth dataset~\cite{pellegrini2009You, lerner2007crowds}. We then use the same trajectory to pass through these three trained models independently and analyze the LN statistics in the first LN layer of the Trajectory Decoder. The input feature of this layer has a dimension of $20 \times 256$, and we chart these values across the 20 dimensions depicted in \cref{fig:ln_decoder}, as the LN affects the last dimension. Our observations reveal minimal statistical variance at different observation lengths, indicating that feature representations extracted from the same trajectory at varying observation lengths have a very similar statistical structure for subsequent decoding (prediction). We conduct further experiments that apply three additional specialized LN layers in the Trajectory Decoder on the nuScenes dataset~\cite{zhan2019interaction}, detailed in \cref{tab:aformer_ln}. Applying specialized LN to the first LN layer results in almost the same performance. The addition of two or three specialized LN layers within the Trajectory Decoder shows minimal improvement. Consequently, we decide to implement specialized LN layers only in the Transformer Encoder to balance performance with model complexity.

The HiVT model uses several LN layers in each of its components. Yet, our observations indicate that a notable statistical discrepancy arises primarily in the Transformer Encoder. We conduct additional experiments that involve the use of additional specialized LN layers in different components on the Argoverse 1~\cite{chang2019argoverse} validation set, as detailed in \cref{tab:hivt_ln}. The findings are consistent with our observations from the AgentFormer model. Consequently, we decide to implement only two specialized LN layers in the Temporal Transformer Encoder.

In conclusion, it becomes evident that normalization shifts typically occur within the Transformer Encoder (Temporal Modeling) when the observation lengths are different. This shift is also one of the reasons for the performance drop. This finding is consistent with the empirical results discussed in the main section of the paper.

\begin{table}[t]
    \centering
    \scalebox{0.75}{
    \begin{tabular}{lcccc}
    \toprule
    \multicolumn{1}{l}{\multirow{2}{*}{Method}} 
    & \multicolumn{1}{c}{\multirow{2}{*}{Note}}
    & \multicolumn{3}{c}{ADE$_{5}$/FDE$_{5}$ $\downarrow$ \quad K = 5 Samples} \\
    \cmidrule(lr){3-5}
    & & 2Ts & 3Ts & 4Ts \\
    \midrule
         AFormer-IT~\cite{yuan2021agentformer} & -
         & 2.02/4.23 
         & 1.93/3.97	
         & 1.86/3.89\\
         \hdashline
         AFormer-FLN & -
         & 1.92/3.91	
         & 1.88/3.89	
         & 1.83/3.78 \\
         \cmidrule(lr){1-5}
         AFormer-FLN 
         & +1 SLN
         & 1.92/3.92	
         & 1.87/3.89	
         & 1.83/3.78 \\
         AFormer-FLN
         & +2 SLN
         & 1.91/3.91	
         & 1.87/3.88	
         & 1.83/3.77 \\
         AFormer-FLN
         & +3 SLN
         & 1.91/3.90	
         & 1.86/3.88
         & 1.82/3.76 \\
        \bottomrule
    \end{tabular}
    }
    \caption{Specialized Layer Normalization study in the AgentFormer model on the nuScenes dataset. The term +$l$ SLN refers to applying $l$ extra Specialized Layer Normalization layers within the Trajectory Decoder.}
    \label{tab:aformer_ln}
\end{table}

\begin{table}[t]
    \centering
    \scalebox{0.75}{
    \begin{tabular}{lcccc}
    \toprule
    \multicolumn{1}{l}{\multirow{2}{*}{Method}} 
    & \multicolumn{1}{c}{\multirow{2}{*}{Note}}
    & \multicolumn{3}{c}{ADE$_{6}$/FDE$_{6}$ $\downarrow$ \quad K = 6 Samples} \\
    \cmidrule(lr){3-5}
    & & 10Ts & 20Ts & 30Ts \\
    \midrule
         HiVT-64-IT~\cite{zhou2022hivt} & -
         & 0.92/1.43
         & 0.81/1.17
         & 0.69/1.04 \\
         \hdashline
         HiVT-64-FLN
         & 
         & 0.81/1.25
         & 0.72/1.08
         & 0.65/0.98 \\
         \cmidrule(lr){1-5}
         HiVT-64-FLN
         & +2 SLN-SE
         & 0.80/1.24	
         & 0.72/1.08	
         & 0.65/0.98  \\
         
         HiVT-64-FLN
         & +4 SLN-TD
         & 0.81/1.24	
         & 0.71/1.07	
         & 0.64/0.97  \\
        \bottomrule
    \end{tabular}
    }
    \caption{Specialized Layer Normalization study in the HiVT-64 model using the Argoverse 1 validation set. The term +2 SLN-SE denotes the addition of 2 Specialized Layer Normalizations in the Agent-Agent Interaction module within the HiVT's Spatial Encoder, while +4 SLN-TD refers to the addition of 4 additional Specialized Layer Normalizations in the Trajectory Decoder.}
    \label{tab:hivt_ln}
\end{table}

\section{Validation of Normalization Shift}
\begin{figure}[t]
\centering
    \begin{subfigure}[b]{0.23\textwidth}
        \centering
        \includegraphics[width=\textwidth]{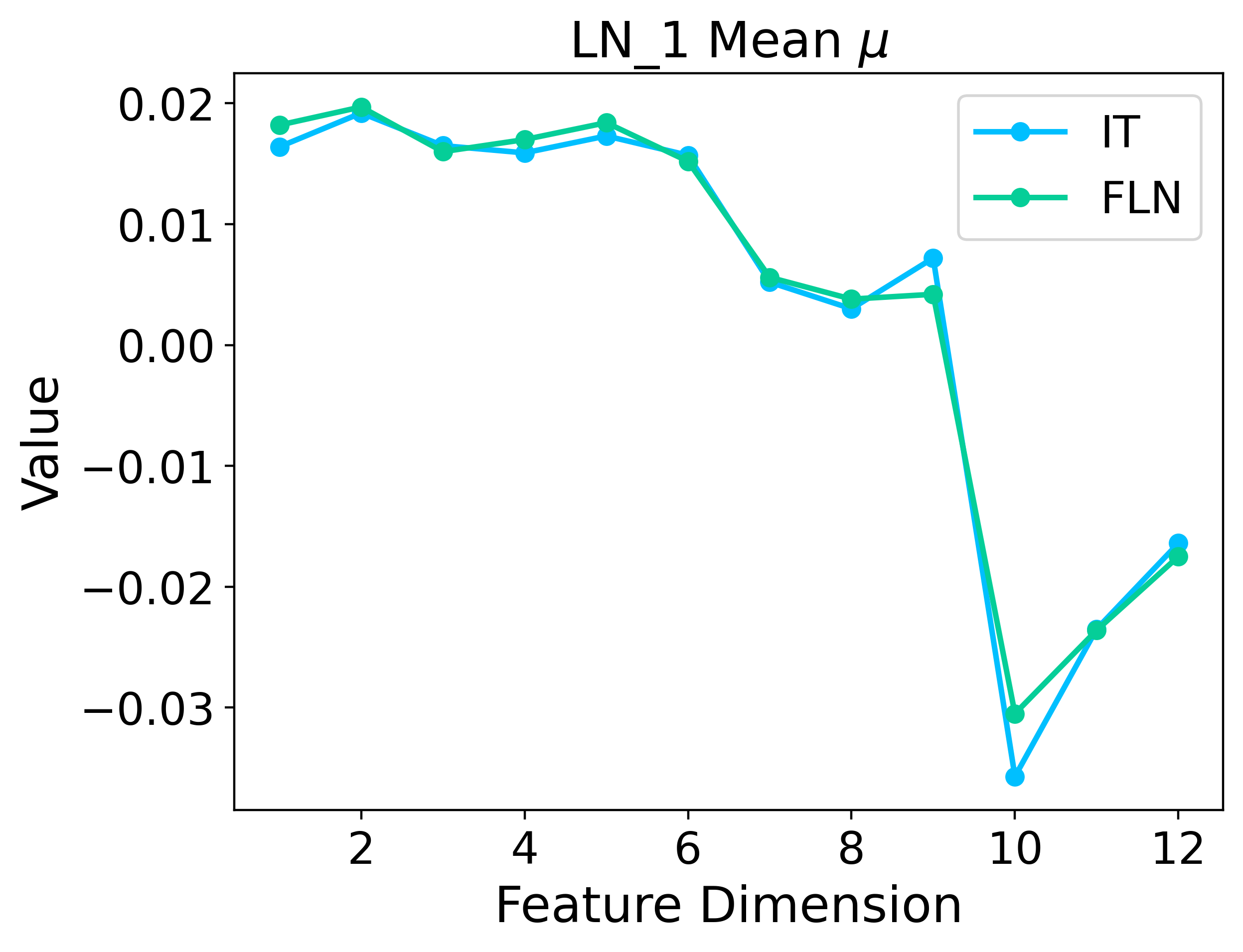}
    \end{subfigure}
    \begin{subfigure}[b]{0.23\textwidth}
        \centering
        \includegraphics[width=\textwidth]{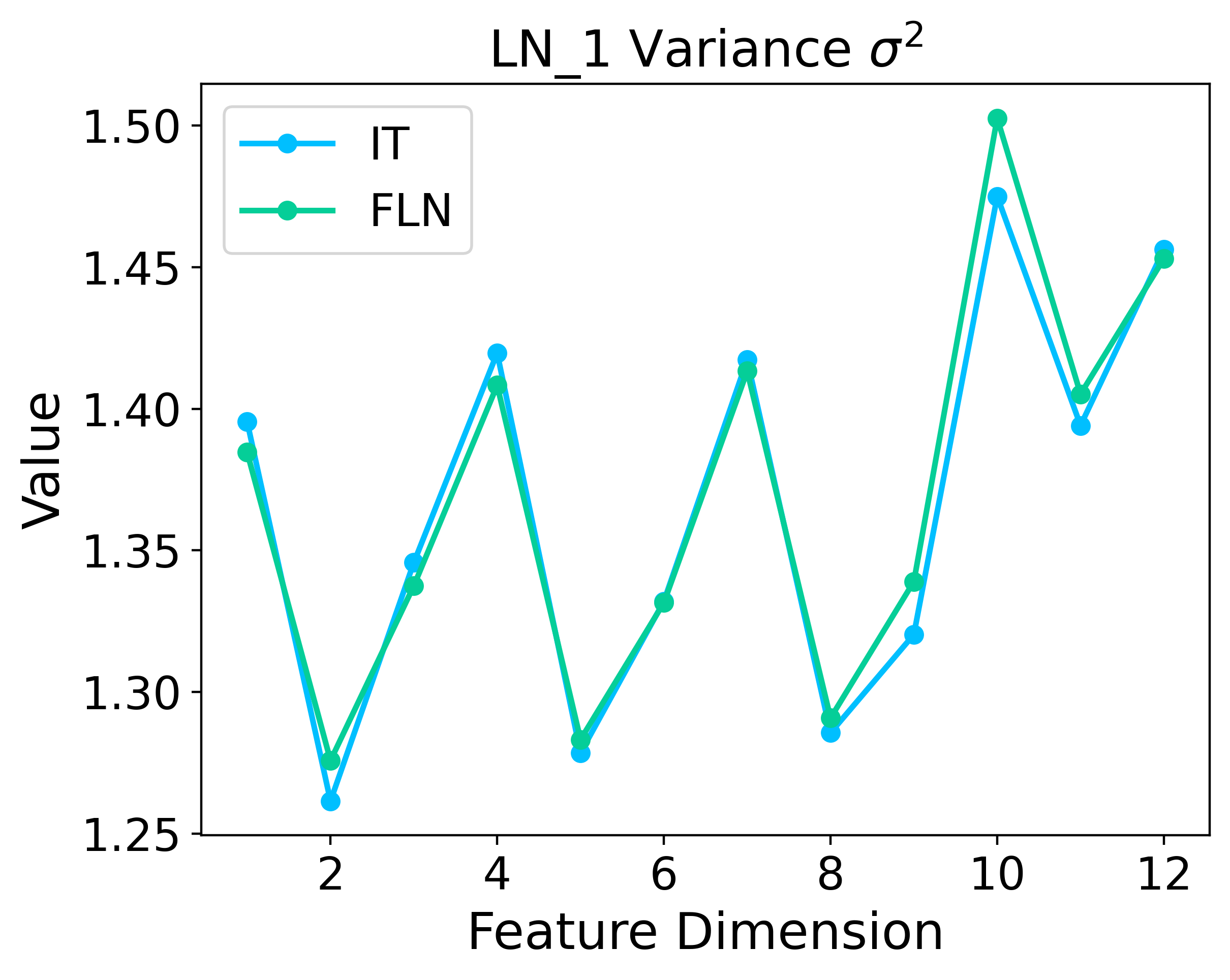}
    \end{subfigure}
    \caption{Layer Normalization statistics for the first layer of the Transformer Encoder in the AgentFormer model, with IT trained at the observation length of 4 timesteps. Both IT and FLN are evaluated at the same observation length of 4 timesteps.}
    \label{fig:ln_trans}
\end{figure}

To further validate that our FlexiLength Network (FLN) can alleviate the Normalization Shift problem, we analyze the Layer Normalization (LN) statistics between Isolated Training (IT) and FLN, both trained using the AgentFormer model on the nuScenes dataset. We use an identical trajectory with 3 agents over an observation period of 4 timesteps and process it through both IT and FLN models separately. The first LN layer in these models receives an intermediate input feature with dimensions of $12 \times 256$. Since LN operates along the last dimension, the statistical outcomes, depicted in \cref{fig:ln_trans}, are presented along the 12-dimensional axis. The alignment of the two curves indicates that the statistical values are quite similar, demonstrating the effectiveness of FLN in mitigating the normalization shift problem.

\section{Quantitative Results}
In the Experiments section, we present the performance of our proposed FlexiLength Network (FLN) on the ETH/UCY dataset through various figures. Additionally, we include corresponding quantitative results in \cref{tab:eth-ucy} for further reference. It is evident that our FLN consistently surpasses the performance of Isolated Training (IT) across different observation lengths in all five datasets.

\begin{table}[t]
\centering
   \scalebox{0.75}{
    \begin{tabular}{lcccc}
    \toprule
     \multicolumn{1}{l}{\multirow{2}{*}{Method}} 
     & \multicolumn{1}{c}{\multirow{2}{*}{Dataset}}
     & \multicolumn{3}{c}{ADE$_{20}$/FDE$_{20}$ $\downarrow$ \quad K = 20 Samples}\\
    \cmidrule(lr){3-5}
     & & 2 Ts & 6 Ts & 8 Ts \\
    \midrule
    AFormer~\cite{yuan2021agentformer} 
    & \multirow{3}{*}{Eth} 
    & 0.661/0.966 
    & 0.640/0.946 
    & 0.451/0.748 \\
    AFormer-IT  
    && 0.467/0.757	
    & 0.452/0.757	
    & 0.451/0.748 \\
    \textbf{AFormer-FLN}
    && \textbf{0.450/0.742}
    & \textbf{0.432/0.730}
    & \textbf{0.411/0.721} \\
    
    \cmidrule(lr){1-5}
    
    AFormer~\cite{yuan2021agentformer} 
    & \multirow{3}{*}{Hotel}
    & 0.225/0.349
    & 0.166/0.277
    & 0.142/0.225 \\
    AFormer-IT 
    && 0.161/0.276
    & 0.148/0.242
    & 0.142/0.225 \\
    \textbf{AFormer-FLN} 
    && \textbf{0.153/0.248}
    & \textbf{0.138/0.232}
    & \textbf{0.124/0.210} \\
    
    \cmidrule(lr){1-5}
    
    AFormer~\cite{yuan2021agentformer}
    & \multirow{3}{*}{Univ}
    & 0.341/0.538
    & 0.275/0.475	
    & 0.254/0.454 \\
    AFormer-IT 
    && 0.263/0.478	
    & 0.251/0.458	
    & 0.254/0.454 \\
    \textbf{AFormer-FLN} 
    && \textbf{0.251/0.457}	
    & \textbf{0.244/0.447}
    & \textbf{0.232/0.430} \\
    
    \cmidrule(lr){1-5}
    
    AFormer~\cite{yuan2021agentformer}
    & \multirow{3}{*}{Zara1}
    & 0.250/0.412	
    & 0.212/0.347	
    & 0.177/0.304 \\
    AFormer-IT 
    && 0.184/0.319	
    & 0.179/0.310	
    & 0.177/0.304 \\
    \textbf{AFormer-FLN} 
    && \textbf{0.178/0.308}	
    & \textbf{0.162/0.300}	
    & \textbf{0.160/0.288} \\
    
    \cmidrule(lr){1-5}
    
    AFormer~\cite{yuan2021agentformer}
    & \multirow{3}{*}{Zara2}
    & 0.190/0.312	
    & 0.178/0.286	
    & 0.140/0.236 \\
    AFormer-IT 
    && 0.140/0.238	
    & 0.142/0.241	
    & 0.140/0.236 \\
    \textbf{AFormer-FLN} 
    && \textbf{0.131/0.230}	
    & \textbf{0.131/0.221}
    & \textbf{0.128/0.217} \\
    \bottomrule
    \end{tabular}
     }
    \caption{Comparison with baseline models on the ETH/UCY dataset, evaluated using the ADE$_{20}$/FDE$_{20}$ metric. The best results are highlighted in bold.}
\label{tab:eth-ucy}
\end{table}

\section{Visualizations}
\begin{figure}[t]
\centering
    \begin{subfigure}[b]{0.15\textwidth}
        \centering
        \includegraphics[width=\textwidth]{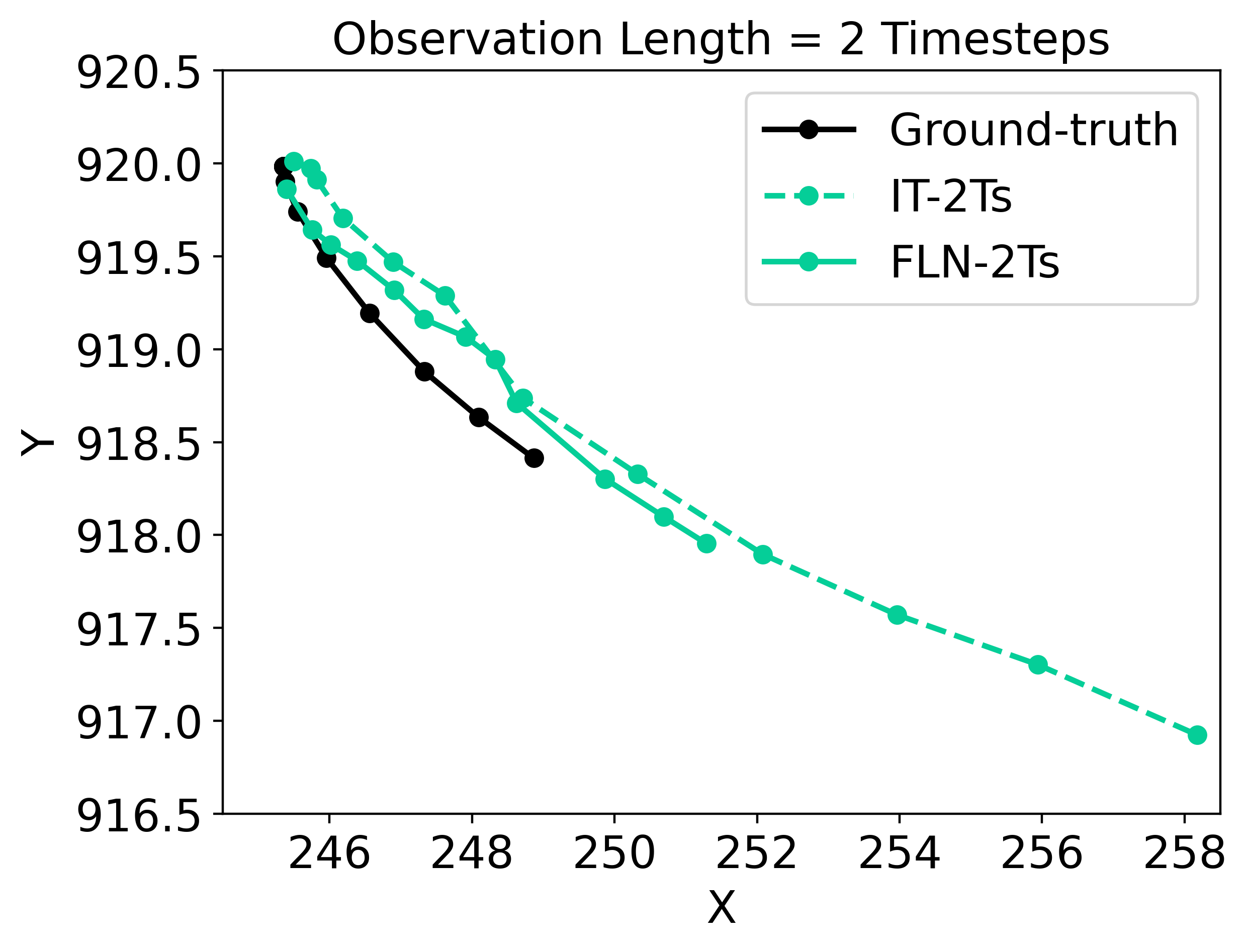}
    \end{subfigure}
    \hspace{-1mm}
    \begin{subfigure}[b]{0.15\textwidth}
        \centering
        \includegraphics[width=\textwidth]{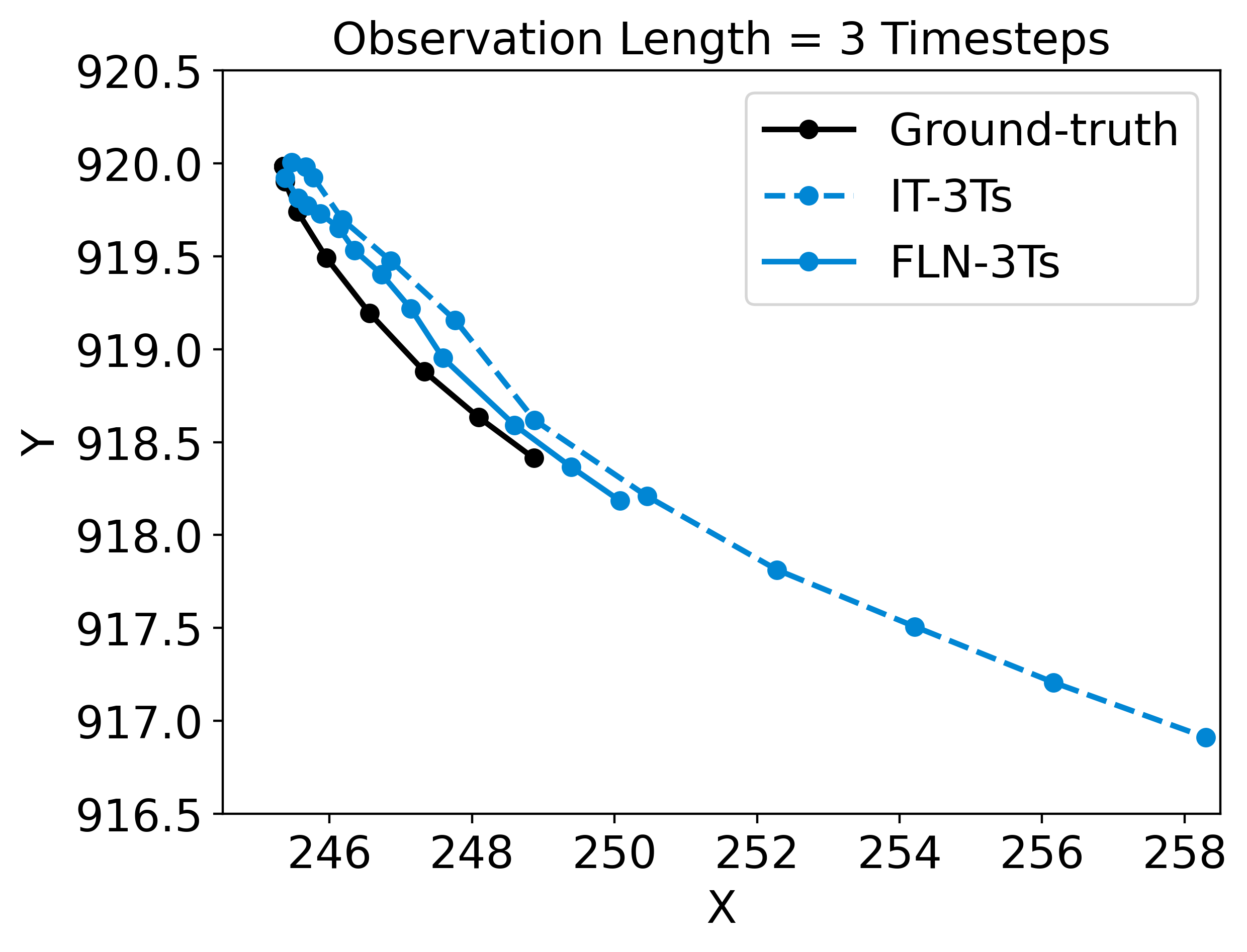}
    \end{subfigure}
    \hspace{-1mm}
    \begin{subfigure}[b]{0.15\textwidth}
        \centering
        \includegraphics[width=\textwidth]{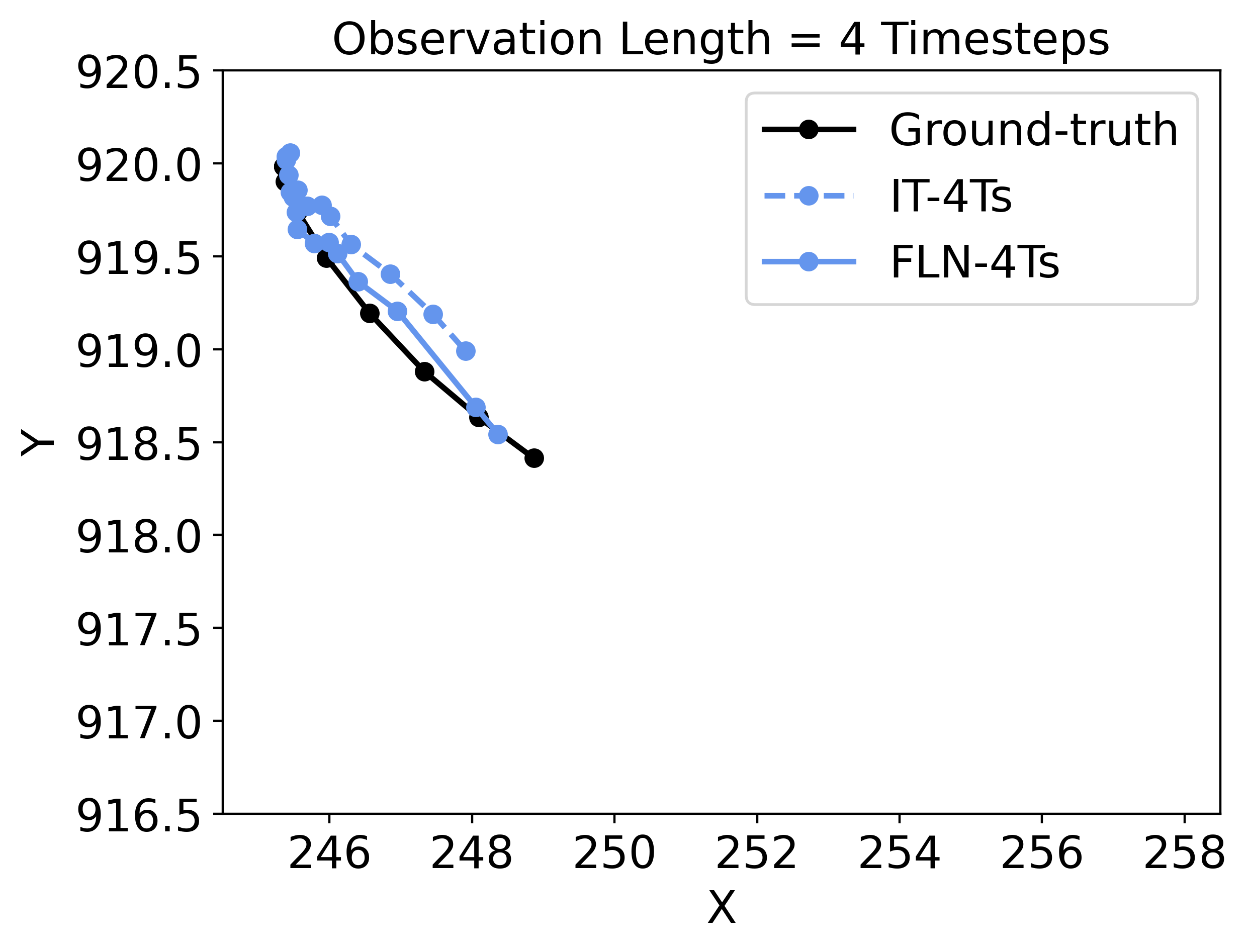}
    \end{subfigure}
    \caption{Visualizations of trajectory predicted by Isolated Training (IT) and Our FlexiLength Network (FLN).}
    \label{fig:vis}
\end{figure}

In \cref{fig:vis}, we present trajectory prediction visualizations using the AgentFormer model on the nuScenes dataset. These visualizations showcase the same trajectory but with different observation lengths. We focus on a single agent and maintain the figure size for easier comparison. These visualizations clearly demonstrate that our FlexiLength Network (FLN) outperforms Isolated Training (IT) across various observation lengths, confirming the effectiveness of FLN in handling inputs with differing observation lengths.

{
    \newpage
    \small
    \bibliographystyle{ieeenat_fullname}
    \bibliography{reference}
}


\end{document}